\documentclass[runningheads]{llncs}

\usepackage{accv}

\usepackage{accvabbrv}

\usepackage{graphicx}
\usepackage{booktabs}

\usepackage{amssymb}
\usepackage{multirow}
\usepackage{pifont}
\newcommand{\xmark}{\ding{53}}
\usepackage{wasysym}
\usepackage{graphicx}
\usepackage{siunitx}

\usepackage[accsupp]{axessibility}  

\usepackage{hyperref}

\usepackage{orcidlink}

\begin{document}

\title{ReLEAP: Upcycling Vision Datasets through Cross-Task Unification and Semi-Supervised Annotation}
\titlerunning{ReLEAP: Upcycling Vision Datasets}

\author{Julia Simon\orcidlink{0000-0003-0243-9706} \and
Daniel Steininger\orcidlink{0000-0003-3810-1803} \and
Andreas Trondl\orcidlink{0009-0006-5748-3607}}
\authorrunning{J.~Simon et al.}

\institute{AIT Austrian Institute of Technology, Giefinggasse 4, 1210 Vienna, Austria\\
Center for Vision, Automation \& Control\\
\email{\tt\small\{julia.simon,daniel.steininger,andreas.trondl.fl\}@ait.ac.at}}

\maketitle
    
\begin{abstract}
Developing robust vision models requires large amounts of annotated data, especially for tasks involving fine‑grained class distinctions under variable environmental conditions. Even in domains with numerous existing datasets, differences in annotation formats, label definitions and targeted learning tasks hinder data reuse and integration. To address this challenge, we introduce \textit{ReLEAP}, a scalable Python framework for unifying heterogeneous image datasets through data‑source abstraction, systematic reuse of annotations across learning tasks, and mapping of labels to a shared hierarchical taxonomy. We demonstrate its applicability in precision agriculture by integrating fifteen diverse datasets covering a broad range of plant species and annotation types. Beyond unification, \textit{ReLEAP} enables systematic recycling of existing data and provides a robust foundation for learning‑based annotation upcycling. Leveraging this foundation, we conduct extensive cross‑dataset and fine‑tuning experiments to investigate how domain‑specific models trained on unified data and foundation models can be used to automatically generate higher‑level annotations under unsupervised, weakly supervised and semi‑supervised settings, and how dataset characteristics influence their effectiveness. Our results show that unified data access and annotation upcycling substantially expand the pool of reusable training data and annotations while improving data transparency, enabling more sustainable and resource‑efficient use of existing datasets. We release the framework and trained models for academic use to support accessible and collaborative research, particularly in data‑scarce and heterogeneous domains.
\end{abstract}

\section{Introduction}
Training broadly applicable vision models typically requires large volumes of annotated data tailored to specific domains and learning tasks. Acquiring this data remains time-consuming and resource-intensive, particularly in niche applications that are not sufficiently covered by state-of-the-art foundation models and require distinction between visually similar classes under diverse environmental conditions. Precision agriculture exemplifies this challenge: detecting and localizing specific plant species requires robust differentiation among similar plant types across varying growth stages, weather conditions and seasonal effects, all of which significantly influence visual appearance. Even with efficient recording pipelines and semi-automatic annotation tools, the efforts and required expert knowledge for fine-grained label assignment of rare species remain substantial. 

\begin{figure}
    \centering
    \includegraphics[width=1\columnwidth]{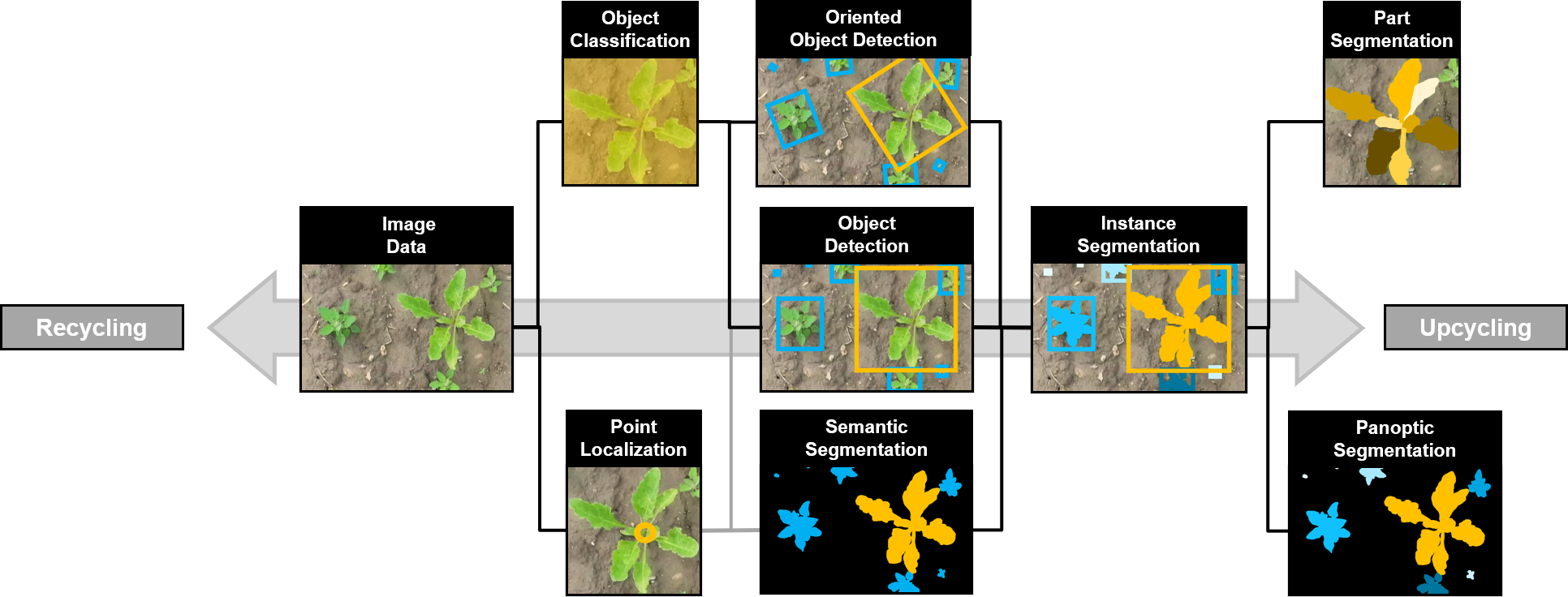}
    \caption{Hierarchy of common learning tasks in computer vision (label colors indicate object categories and hues of the same color individual objects). While all annotations can potentially be upcycled to higher-level tasks using specialized models, only those connected by black lines are suitable for direct recycling to the next lower level.}
    \label{fig:tasks_overview}
\end{figure}

A more sustainable alternative is to leverage existing annotated datasets from public sources. However, even if they are available in a given domain, no single one typically captures the variability required for a novel application scenario. On the other hand, integrating multiple datasets into a unified training corpus is usually nontrivial due to heterogeneous recording protocols, annotation formats, label definitions and targeted learning tasks, as visualized in \cref{fig:tasks_overview}.

This fragmentation often results in insufficient annotated data for specific combinations of tasks and label classes, as well as additional efforts for suitability assessment and conversion of available data. An efficient approach to analyzing and fusing diverse datasets across learning tasks could unlock the latent diversity of existing resources. Even greater potential lies in automatically extending the applicability of existing data to new tasks and label spaces. In combination, these innovations could reduce the effort required for data collection, annotation and aggregation in novel scenarios, thereby presenting a relevant step towards more sustainable model training. We propose the following key contributions:

\begin{enumerate}
    \setlength\itemsep{0.55em}
    \item
        We introduce \textit{ReLEAP},\footnote{Available for academic use at
        \href{https://github.com/releap-vision/releap}{https://github.com/releap-vision/releap}} a versatile framework for unifying and analyzing heterogeneous image datasets and reusing annotations across learning tasks and label spaces, demonstrated in precision-agriculture.
    \item
        We systematically evaluate strategies for annotation upcycling and automatic labeling under varying supervision, model choices, dataset characteristics and domain shifts through cross-dataset and fine-tuning experiments.
    \item
        Based on an extensive survey, we integrate 15 diverse datasets into \textit{ReLEAP} and release our framework and trained models to support accessible and sustainable vision research.
\end{enumerate}
\section{Related Work}
\label{sec:related_work}
Integrating heterogeneous datasets for vision tasks has been explored mostly in the context of domain adaptation and label harmonization. One common approach is to adapt model architectures to handle inconsistencies in annotation policies and granularities across datasets for different tasks and domains \cite{LiAc24,ZhKo22,XuFa20,RaLa22,ZhSc20}. Other works focus on unifying label spaces on the level of annotations by applying hand-crafted or learned ontologies \cite{KeAv25,CrNi21,LaLi20}, a direction that is complementary to our approach, as we provide means to efficiently integrate these taxonomies for label harmonization and annotation upcycling across tasks.

Repurposing existing annotations for new tasks has gained traction as a cost-effective alternative to manual labeling. Papadopoulos et al. \cite{PaWe21} introduced label propagation for scaling instance segmentation, while Kennerley et al. \cite{KeAv25} proposed Label-Aligned Transfer to unify object-detection datasets. Liao et al. \cite{LiAc24} extended this concept with taxonomy-guided pseudo-labeling. These methods show promising results but are typically limited to single-task adaptation.

Semi-supervised learning reduces annotation costs by leveraging unlabeled data \cite{LiKa21,ReEh20}, often in the context of active-learning strategies \cite{PaWe21,JaGr16}. Teacher-student frameworks and adaptive pseudo-labeling dominate this field \cite{ZhYa23,HoDa24, WaWe24}. While effective, they operate within single datasets and do not address cross-dataset annotation scaling.

While prior research addresses dataset harmonization, semi-supervised annotation and cross-dataset generalization, few approaches systematically combine these aspects into a unified pipeline for dataset unification and annotation upcycling across tasks (a detailed feature-level comparison is provided in the supplementary material). Our work fills this gap by proposing a generic framework that integrates harmonization, taxonomy-based label mapping and systematic recycling of annotations across tasks, combined with scientifically evaluated upcycling strategies to enable scalable reuse of existing datasets.
\section{Data Unification and Recycling}
\label{sec:recycling}
We introduce \textit{ReLEAP} (Reusing vision data through Label harmonization and Efficient Annotation Propagation), a scalable framework for analyzing and unifying heterogeneous image datasets and extending their utility across vision tasks. To maximize accessibility, it is implemented as an open-source Python project and designed to provide a scalable structure, easily extensible to new datasets.

\subsection{Framework Design}
\begin{figure}
    \centering
    \includegraphics[width=1\columnwidth]{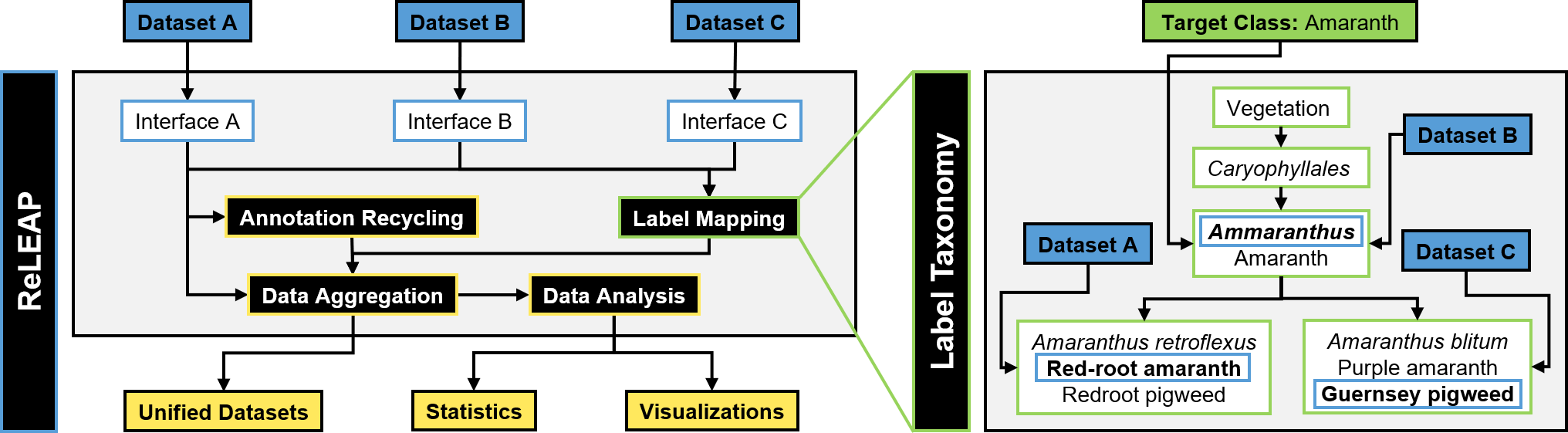}
    \caption{\textit{ReLEAP} framework structure and main modules, along with an example for label mapping using a unified taxonomy. Each taxonomy node defines synonymic names and parent classes for a category. Automatically assigning both source and target class names to these nodes allows aggregation across data sources at arbitrary granularity.}
    \label{fig:schematic_overview}
\end{figure}

As illustrated in \cref{fig:schematic_overview}, \textit{ReLEAP} abstracts data sources, allowing all core functionality to operate independently of the underlying dataset structure. Each dataset only requires a lightweight interface specifying methods for retrieving images, annotations and metadata. Shared modules handle interactions with common hosting platforms (e.g., Zenodo \cite{Ze13}, Roboflow \cite{DwNe26}, Google Drive) and conversions from standard annotation formats. This separation minimizes dataset-specific code and enables efficient integration of new sources. The core modules described below then operate independently of the underlying data source.

\subsubsection{Taxonomy-Based Label Mapping.} A key component of \textit{ReLEAP} is its automated mapping of heterogeneous labels to arbitrary target label spaces using a shared taxonomy. Datasets often describe identical concepts using different names, granularities or even languages. While some of these equivalences are trivial to identify, others are more obscure and require significant domain knowledge. To address this, we employ a hierarchical taxonomy encoding distilled expert knowledge, which is defined once per application domain and reused across datasets. By mapping both source and target labels to its nodes, sources can be assigned to the next connected target of higher order, enabling consistent alignment of labels across datasets and aggregation at arbitrary levels of detail. For precision agriculture, the taxonomy was derived from botanical ontologies \cite{Wi26, Ep26} defining a hierarchy from plant families to individual species, along with synonym mappings. For other domains, existing ontologies may provide suitable taxonomies, while LLM-assisted generation could potentially reduce the effort required to establish an initial hierarchy.

\subsubsection{Data Aggregation.} \textit{ReLEAP} currently supports annotations for classification, detection, semantic and instance segmentation and the agriculture‑specific tasks of stem localization and leaf separation. Configurable aggregation modules automatically select datasets relevant to a target task and label set and download them on demand. To ensure reproducibility, each image receives a unique cached identifier. Annotations are subsequently filtered according to user-defined constraints such as object size, included labels and handling of unlabeled images. The resulting multi-source dataset can then be exported in common annotation formats with configurable image pre-processing and train-val-test splits.

\subsubsection{Annotation Recycling.} Beyond unification, \textit{ReLEAP} systematically extends existing datasets to additional learning tasks through deterministic annotation conversions. Examples include deriving bounding boxes from instance masks or cropping classification samples from detection data. While individually simple, integrating these transformations directly into the aggregation pipeline allows us to automatically maximize the amount of usable data for a given application.

\subsubsection{Analysis and Visualization.} \textit{ReLEAP} further provides configurable modules for jointly analyzing and visualizing original and recycled annotations from multiple sources. Available outputs include image and instance statistics, label distributions before and after filtering, identification of invalid annotations and unlabeled images, and visual inspection of annotated samples.

~\\
By combining these functionalities into an accessible, highly configurable open-source framework, \textit{ReLEAP} streamlines the creation of balanced unified datasets and reduces manual data-preparation effort in heterogeneous domains.

\subsection{Data Integration and Analysis}
\label{sec:survey}
To demonstrate \textit{ReLEAP}’s capabilities, we integrated interfaces for 15 publicly available precision‑agriculture datasets, selected based on an extensive survey of image and annotation quality, diversity of environmental conditions, accessibility and coverage of relevant learning tasks. The dataset collection spans major crops (\textit{maize}, \textit{sugar beet}, \textit{multiple bean species}), common weeds (e.g., \textit{goosefoot}, \textit{thistle}, \textit{knotweed}, \textit{amaranth}, \textit{grasses}) and several niche species to support generalization studies. \cref{tab:stats_selected_datasets} summarizes key statistics inferred by our framework, including image and instance counts, supported tasks and plant species. More details and statistics of all surveyed datasets are shown in the supplementary material.

Apart from covering sufficient variability of explicitly annotated learning tasks as well as labeling granularity and image sizes, the combined datasets exhibit wide variation in viewpoints, environmental conditions and annotation schemes. While the suitability for rule-based recycling of annotations depends solely on the annotated learning task (see \cref{fig:tasks_overview}), learning-based upcycling is also related to the characteristics of each dataset (\eg classification images can only be used for detection tasks if they are not tightly cropped to the depicted instances). Rule‑based recycling alone increases usable samples substantially, for example, by factors of 1.3 for detection and 27 for classification, while expanding the number of label classes by a factor of 4.2 for the latter. Additional gains can be achieved through upcycling to higher‑level tasks, further unlocking the latent potential of existing datasets and their seamless combination.

\begin{table}[t]
    \caption{Statistical overview of datasets integrated in \textit{ReLEAP} and their suitability for the tasks of \textbf{Cl}a\textbf{s}sification, Object \textbf{Det}ection, Semantic \textbf{Seg}mentation, \textbf{I}nstance \textbf{Seg}mentation, \textbf{Stem}-Point Localization and \textbf{Leaf} Segmentation ($\CIRCLE$: explicitly annotated in the original dataset, $\downarrow$: available through automatic recycling, $\uparrow$: available through learning-based upcycling approaches), along with numbers of \textbf{Im}a\textbf{g}es, annotated plant \textbf{Inst}ances (both in thousands) and \textbf{Cat}egories, as well as median image \textbf{Width} and \textbf{Height} in pixels and average image \textbf{Area} in megapixels.}
    \label{tab:stats_selected_datasets}
    \centering
    \resizebox{1\linewidth}{!}{
    \setlength{\tabcolsep}{2pt}
    \begin{tabular}{@{}l@{\quad}cccccc@{\quad}rrrrrr@{}}
        \toprule
        \textbf{Dataset}
            & \rotatebox{90}{\textbf{Cls}}
            & \rotatebox{90}{\textbf{Det}}
            & \rotatebox{90}{\textbf{Seg}}
            & \rotatebox{90}{\textbf{ISeg}}
            & \rotatebox{90}{\textbf{Stem}}
            & \rotatebox{90}{\textbf{Leaf}}
            & \textbf{Img}
            & \textbf{Inst}
            & \textbf{Cat}
            & \textbf{Width}
            & \textbf{Height}
            & \textbf{Area} \\
        \midrule
        \textit{CottonWeedID15} \cite{ChLu22} & $\CIRCLE$ & $\uparrow$ & $\uparrow$ & $\uparrow$ & $\uparrow$ & $\uparrow$ & \multicolumn{2}{c}{5.2} & 15 & 2,789 & 2,800 & 7.1 \\
        \textit{EarlyCropWeed} \cite{EsBo20} & $\CIRCLE$ & $\uparrow$ & $\uparrow$ & $\uparrow$ & $\uparrow$ & $\uparrow$ & \multicolumn{2}{c}{0.5} & 4 & 4,256 & 2,832 & 15.4 \\
        \textit{PlantSeedlings} \cite{GiMo17} & $\CIRCLE$ & - & $\uparrow$ & $\uparrow$ & $\uparrow$ & $\uparrow$ & \multicolumn{2}{c}{5.5} & 12 & 266 & 266 & 0.2 \\
        \midrule
        \textit{SugarbeetsAndWeeds} \cite{Wa21} & $\downarrow$ & $\CIRCLE$ & $\uparrow$ & $\uparrow$ & $\uparrow$ & $\uparrow$ & 0.1 & 1.0 & 2 & 512 & 512 & 0.3 \\
        \textit{LincolnBeet} \cite{SaDa22} & $\downarrow$ & $\CIRCLE$ & $\uparrow$ & $\uparrow$ & $\uparrow$ & $\uparrow$ & 4.4 & 39.2 & 2 & 1,920 & 1,080 & 2.1 \\
        \textit{PakistaniWeed} \cite{Ha23} & $\downarrow$ & $\CIRCLE$ & $\uparrow$ & $\uparrow$ & $\uparrow$ & $\uparrow$ & 9.2 & 42.8 & 7 & 1,280 & 720 & 0.9 \\
        \textit{VegetableCrops} \cite{LaKe22} & $\downarrow$ & $\CIRCLE$ & $\uparrow$ & $\uparrow$ & $\CIRCLE$ & $\uparrow$ & 2.7 & 9.0 & 3 & 2,048 & 1,536 & 2.4 \\
        \textit{WeedCrop} \cite{UpMa25} & $\downarrow$ & $\CIRCLE$ & $\uparrow$ & $\uparrow$ & $\uparrow$ & $\uparrow$ & 1.1 & 17.7 & 13 & 4,579 & 4,640 & 22.3 \\
        \midrule
        \textit{Sunflower} \cite{PoIm24} & - & - & $\CIRCLE$ & $\uparrow$ & $\uparrow$ & $\uparrow$ & 0.5 & - & 2 & 1,296 & 964 & 1.2 \\
        \textit{Sugarbeets2016} \cite{ChLo17} & - & - & $\CIRCLE$ & $\uparrow$ & $\uparrow$ & $\uparrow$ & 12.3 & - & 4 & 1,296 & 966 & 1.3 \\
        \midrule
        \textit{CropAndWeed} \cite{StTr23} & $\downarrow$ & $\CIRCLE$ & $\CIRCLE$ & $\CIRCLE$ & $\CIRCLE$ & $\uparrow$ & 8.0 & 109.8 & 75 & 1,920 & 1,088 & 2.1 \\
        \textit{LettuceMOTS} \cite{HuSu24} & $\downarrow$ & $\downarrow$ & $\downarrow$ & $\CIRCLE$ & $\uparrow$ & $\uparrow$ & 1.3 & 18.4 & 1 & 810 & 1,080 & 0.9 \\
        \textit{SorghumWeedSeg} \cite{JuTh24} & $\downarrow$ & $\CIRCLE$ & $\downarrow$ & $\CIRCLE$ & $\uparrow$ & $\uparrow$ & 0.3 & 5.1 & 3 & 6,000 & 4,000 & 24.0 \\
        \textit{WeedsGalore} \cite{CeKu25} & $\downarrow$ & $\downarrow$ & $\CIRCLE$ & $\CIRCLE$ & $\uparrow$ & $\uparrow$ & 0.2 & 12.8 & 5 & 600 & 600 & 0.4 \\
        \textit{PhenoBench} \cite{WeMa24} & $\downarrow$ & $\downarrow$ & $\CIRCLE$ & $\CIRCLE$ & $\CIRCLE$ & $\CIRCLE$ & 2.2 & 37.4 & 3 & 1,024 & 1,024 & 1.0 \\
        \bottomrule
    \end{tabular}
    }
\end{table}
\section{Data Upcycling}
\label{sec:upcycling}
The most challenging approach to leveraging existing data and annotations is their upcycling to higher-level tasks (see \cref{fig:tasks_overview}). Unlike recycling, which relies on deterministic annotation conversions, upcycling requires learning-based methods to generate annotations that are not directly available from existing data. In this section, we investigate how annotations can be generated for previously unsupported datasets and novel image data under varying degrees of supervision.

We focus on instance segmentation as a representative high-level task, expecting similar trends for less complex annotation types. The goal is to generate accurate object masks and species labels either directly from raw images or by exploiting existing lower-level annotations such as bounding boxes or stem locations. Our focus is automatic annotation, i.e. minimizing required prior information and manual post-processing, rather than real-time inference performance.

We evaluate several models trained on unified data provided by \textit{ReLEAP} and compare them with approaches based on SAM 3 \cite{CaGu25}, as summarized in \cref{fig:labeling_variants}. Additional implementation details are provided in the supplementary material.

\begin{figure}[t]
    \centering
    \includegraphics[width=0.65\columnwidth]{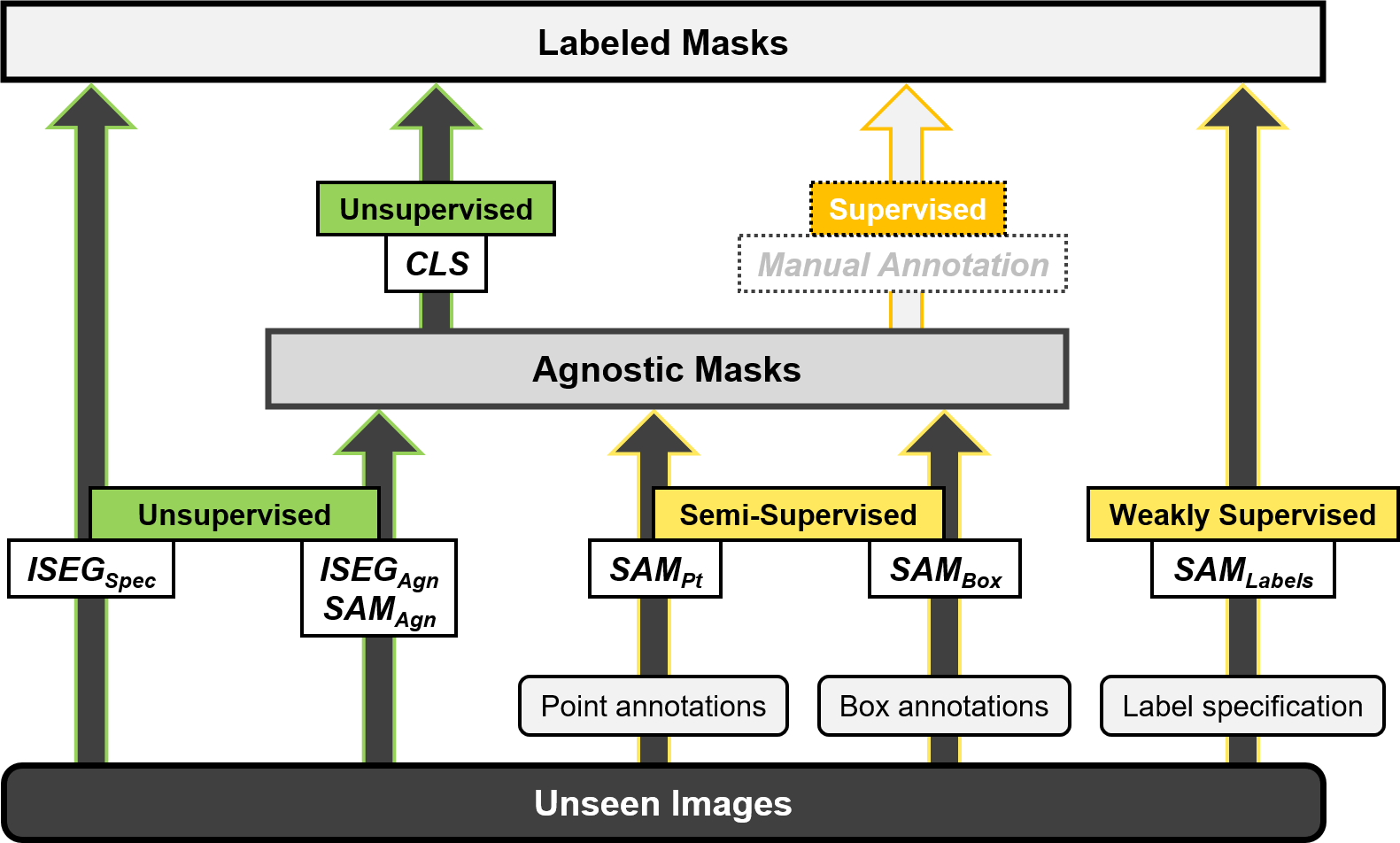}
    \caption{Evaluated annotation-upcycling methods inferring instance-segmentation masks from unannotated image data and varying degrees of prior information (rounded boxes). \textit{ISEG} and \textit{CLS} refer to models trained on instance-segmentation and classification data aggregated by \textit{ReLEAP}. \textit{SAM} denotes a SAM 3 foundation model prompted with the priors in subscript. \textit{Spec} and \textit{Agn} identify models trained on all available classes individually or mapped to a single superclass.}
    \label{fig:labeling_variants}
\end{figure}

\subsubsection{Unsupervised Methods.}
These methods create new instance masks without prior information beyond the input images. A straightforward approach is to train a single multi-class model on the combined data of all available \textit{ReLEAP} datasets containing instance-segmentation annotations. Dataset labels are aligned through the unified taxonomy, yielding a model capable of directly predicting species-specific instance masks (\textit{ISEG\textsubscript{Spec}}). While this approach benefits from diverse training sources for accurate delineation, its label space is limited by the classes available in instance-segmentation datasets.

Class coverage can be increased by separating mask generation and class assignment into two processing stages. Class-agnostic instance masks for a shared superclass (e.g. \textit{vegetation}) can be provided by a single-class model trained on all \textit{ReLEAP} instance-segmentation datasets (\textit{ISEG\textsubscript{Agn}}). Alternatively, SAM 3 is prompted with the names \textit{plant}, \textit{vegetation}, \textit{crop} and \textit{weed} and the resulting masks are merged into a common class using non-maxima suppression (\textit{SAM\textsubscript{Agn}}). Fine-grained species assignment for each detected instance is then performed by a separate classifier (\textit{CLS}) trained on all classification data available through \textit{ReLEAP}. This classifier benefits from the considerably larger pool of classification annotations, including additional samples generated through recycling.

\subsubsection{Weakly Supervised and Semi-Supervised Methods.}
The annotation task becomes easier when prior information is available. If only the expected object categories are known, their names can be used as text prompts for SAM 3 to directly generate labeled instance masks (\textit{SAM\textsubscript{Labels}}). When location information is available in the form of bounding boxes or stem points, deriving instance masks is further simplified. Outputs from \textit{ISEG\textsubscript{Agn}} or \textit{SAM\textsubscript{Agn}} can be matched to existing boxes through IoU-based filtering (\textit{ISEG\textsubscript{Agn}}+\textit{Box}, \textit{SAM\textsubscript{Agn}}+\textit{Box}), replacing the need for a classifier. Alternatively, SAM 3 can be prompted directly with boxes or points to produce instance masks (\textit{SAM\textsubscript{Box}}, \textit{SAM\textsubscript{Pt}}).

\subsubsection{Few-Shot Domain Adaptation.}
In addition to comparing the proposed upcycling strategies, we perform few-shot experiments to evaluate how efficiently unsupervised methods can be adapted to new domains. Specifically, we fine-tune them using varying numbers of manually annotated target samples and measure the resulting improvements in automatic-annotation quality.

\subsection{Evaluation Protocol}
\label{sec:evaluation_protocol}
Since the objective of upcycling is to automatically annotate previously unseen data under varying levels of supervision, our evaluation protocol simulates the corresponding domain shift using the five instance-segmentation datasets integrated in \textit{ReLEAP} (see \cref{tab:stats_selected_datasets}). We conduct five evaluation rounds, each time treating one dataset as the target set to be annotated while excluding it entirely from model training. All upcycling methods are applied to the unseen target data and their results compared to the original manual annotations. The aim is not to benchmark segmentation models for deployment, but to assess how closely automatically generated annotations match human-created instance masks.

As a reference, we additionally train a model on each instance-segmentation dataset individually and evaluate its transfer to the others. This baseline represents a common strategy for semi-automatic annotation and allows us to quantify the benefit of multi-source aggregation relative to single-dataset training.

\subsubsection{Metrics.}
We use the challenging \textit{mAP\textsuperscript{50-95}} metric \cite{LiMa14} for comparing segmented masks on each target dataset to the corresponding ground truth as an overall measure of how well our automatically generated pre-annotations align with manually created ones. While annotation quality does not directly measure annotation effort, precision and recall are closely related to the number of corrections required during manual post-processing. We therefore report all three metrics (as reported by Ultralytics \cite{JoQi24}) to assess both annotation fidelity and expected correction effort. Since class granularity is heterogeneous across test datasets, we train our class-specific models on the most fine-grained labels available and apply our taxonomy to match them to the potentially more coarse-grained annotations of each target dataset, thereby counting subclasses of an originally annotated class as correct assignments.

We additionally report average inference times for each method to compare efficiency and accessibility. Timings include all processing by our own or foundational models, as well as pre- and post-processing for inference, but exclude times for generating prompts or assigning labels from existing annotations.

\subsubsection{Experimental Setup.}
\label{sec:setup}
All models are trained on a single NVIDIA RTX 4090 and evaluated on an NVIDIA RTX 4070 Ti Super. Prior to training \textbf{instance-segmentation} models, we use \textit{ReLEAP} to resize all images to a width of 1280 pixels, preserving aspect ratios, map classes to a common label space and exclude object instances with mask areas smaller than 96 pixels. For the multi-class model \textit{ISEG\textsubscript{Spec}}, classes providing fewer than 200 instances or present in fewer than 10 images across all five datasets are remapped to parent classes, if possible, or removed along with all images that contain them. For instance-segmentation training, we rely on YOLO26 \cite{JoQi24} with default hyper-parameters and data-augmentation techniques. After initial ablation experiments, we select the \textit{Large} architecture pretrained on COCO \cite{LiMa14} to ensure sufficient capacity for learning features across multiple large datasets. Training is conducted with an input resolution of 896 pixels and adaptive batch size. For each model, the images of four datasets are split into training and validation sets with a ratio of 8:2 individually for the baseline models, which are then combined for the upcycling models, while the test split always consists of the entire remaining dataset. The models with best validation results are selected after training for 100 epochs.

Similarly, for the \textit{CLS} \textbf{classification} models, we combine data from all 13 \textit{ReLEAP} datasets providing classification annotations, either explicitly or by recycling from detection or instance segmentation, except for recycled annotations of the respective target set. The resulting image patches of individual plants are filtered by a minimum original width and height of 16 pixels. We remove conflicting classes based on our hierarchical taxonomy by excluding instances labeled with classes corresponding to higher botanical levels than subfamilia to avoid the inclusion of generic labels like \textit{vegetation}, \textit{crop}, \textit{weed} or \textit{grasses}. Subsequently, we eliminate classes in the target label space containing fewer than 100 instances. This results in a total of 88 available classes comprising 203k instances across all datasets, which are resized to a square size of 192 pixels for training, as this is the power of 32 closest to the mean image patch size of 244.7$\times$186.7 pixels. Classification models are trained for 50 epochs using the MambaOut \cite{YuWa25} \textit{Kobe} architecture, as provided by Pytorch Image Models \cite{Ro19}, with a batch size of 96 and initial learning rate 0.004. Detailed lists of training labels and their distributions for both learning tasks can be found in the supplementary material.

\subsection{Method Comparison}
\label{sec:method_ablation}

We evaluate our upcycling methods on each target dataset as described in \cref{sec:evaluation_protocol}, starting with class-agnostic mask generation and then comparing class-specific models with multiple degrees of supervision. 

\subsubsection{Class-Agnostic Mask Generation.}
We first analyze unsupervised class‑agnostic mask generation with \textit{ISEG\textsubscript{Agn}} and \textit{SAM\textsubscript{Agn}}, and compare the results with a class-agnostic evaluation of our baseline models, as summarized in \cref{tab:results_agnostic}. Both methods consistently outperform the baseline for this task. Results further show that a domain‑specific model trained on multiple unified datasets produces slightly more accurate contours for the \textit{vegetation} superclass, particularly for datasets with larger plant structures, such as \textit{PhenoBench} and \textit{SorghumWeedSeg}. Precision and recall indicate that \textit{ISEG\textsubscript{Agn}} yields more precise contours, while SAM 3 detects more instances overall. The main difference, however, lies in efficiency. The more lightweight specifically trained model outperforms the foundation model by a factor of more than 14 regarding inference time. 

\begin{table}[t]
    \caption{Instance-mask accuracy (\% mAP\textsuperscript{50-95}) of class-agnostic methods by target set (\textbf{P}heno\textbf{B}ench, \textbf{W}eeds\textbf{G}alore, \textbf{S}orghum\textbf{W}eedSeg, \textbf{L}ettuce\textbf{M}OTS, \textbf{C}ropAnd\textbf{W}eed) as well as \textbf{P}recision, \textbf{R}ecall and inference \textbf{t}ime per image, including average and maximum class-agnostic results of baseline models trained on single datasets.}
    \label{tab:results_agnostic}
    \centering
    \resizebox{0.95\linewidth}{!}{
    \begin{tabular}{@{}l@{\qquad}r@{\hspace{0.5em}}r@{\hspace{0.5em}}r@{\hspace{0.5em}}r@{\hspace{0.5em}}r@{\hspace{0.5em}}r@{\qquad}r@{\hspace{0.5em}}r@{\quad}r@{}}
        \toprule
        \textbf{Method} & \multicolumn{1}{l@{\quad}}{\textbf{PB}} & \textbf{WG} & \textbf{SW} & \textbf{LM} & \textbf{CW} & \textbf{Overall} & \multicolumn{1}{@{\hspace{0.5em}}l}{\textbf{P}} & \multicolumn{1}{@{\hspace{0.5em}}l}{\textbf{R}} & \multicolumn{1}{@{\quad}l@{}}{\textbf{t [ms]}} \\
        \midrule
        \textit{Baseline\textsubscript{Overall}} & 27.2 & 17.7 & 12.9 & 57.3 & 21.5 & 27.3 & 69.1 & 34.9 & 34.6 \\
        \textit{Baseline\textsubscript{Best}} & 45.4 & 20.5 & 18.2 & 82.4 & 29.6 & 39.2 & 83.1 & 50.4 & \textbf{31.4} \\
        \textit{SAM\textsubscript{Agn}} & 45.3 & \textbf{22.9} & 16.6 & 85.6 & \textbf{38.8} & 41.9 & 76.4 & \textbf{68.8} & 682.9 \\
        \textit{ISEG\textsubscript{Agn}} & \textbf{54.9} & 18.9 & \textbf{23.3} & \textbf{86.8} & 34.3 & \textbf{43.6} & \textbf{79.0} & 60.9 & 48.2 \\
        \bottomrule
    \end{tabular}
    }
\end{table}

\subsubsection{Unsupervised and Weakly Supervised Approaches.}
\begin{figure}
    \centering
    \includegraphics[width=0.95\columnwidth]{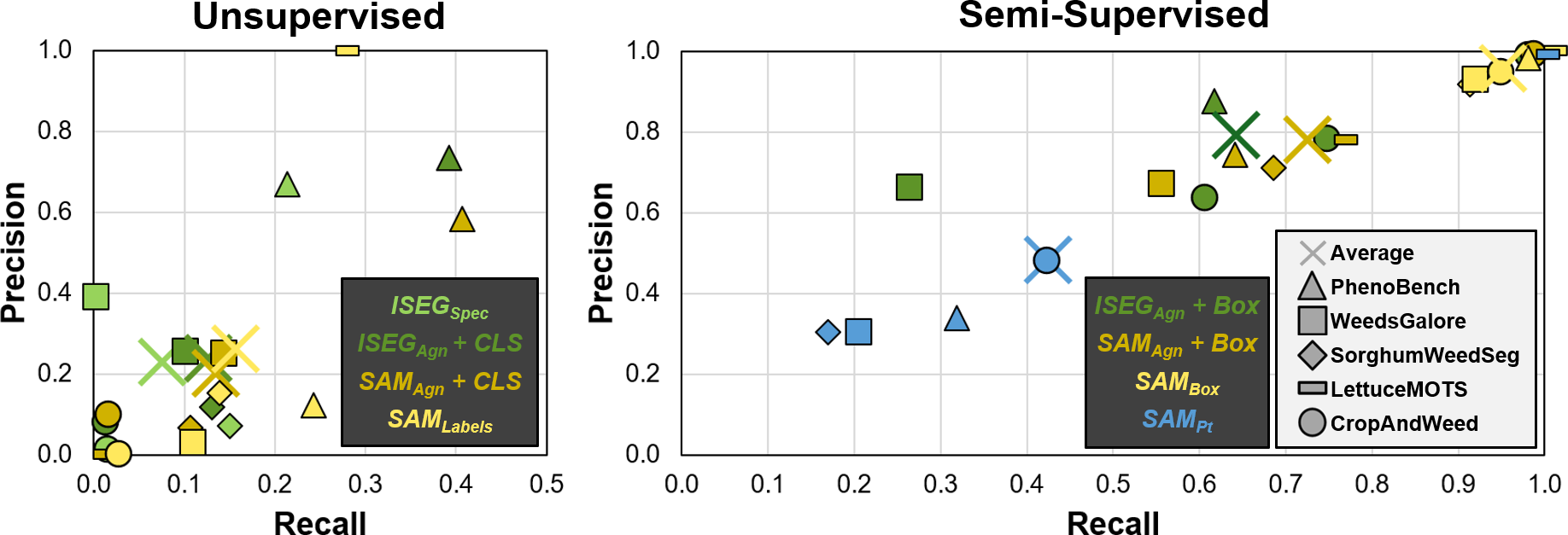}
    \caption{Precision and recall of upcycling approaches applied to each target dataset and their averages. \textit{ISEG} and \textit{CLS} refer to models trained on aggregated \textit{ReLEAP} data for instance segmentation and classification, \textit{SAM} to SAM 3 prompted with priors in subscript. \textit{Spec} and \textit{Agn} denote multi-class and single-superclass models, respectively.}
    \label{fig:methods_pr}
\end{figure}

As illustrated by the precision and recall metrics in \cref{fig:methods_pr}, using domain‑specific models trained on fine-grained classes from multiple sources also leads to competitive performance (\textit{ISEG\textsubscript{Spec}}). While this approach might already be sufficient for use cases with common plant species, even finer distinctions can be achieved with a separate classification stage trained on a larger pool of labels and annotated samples (\textit{ISEG\textsubscript{Agn}}+\textit{CLS}). While both these methods operate without prior information about the data to be annotated, SAM 3 always requires explicit supervision, at least in the form of class prompts (\textit{SAM\textsubscript{Labels}}), or a domain-specific classification model (\textit{SAM\textsubscript{Agn}}+\textit{CLS}) to provide labeled instance masks.

\begin{figure}[t]
    \centering
    \includegraphics[width=1\columnwidth]{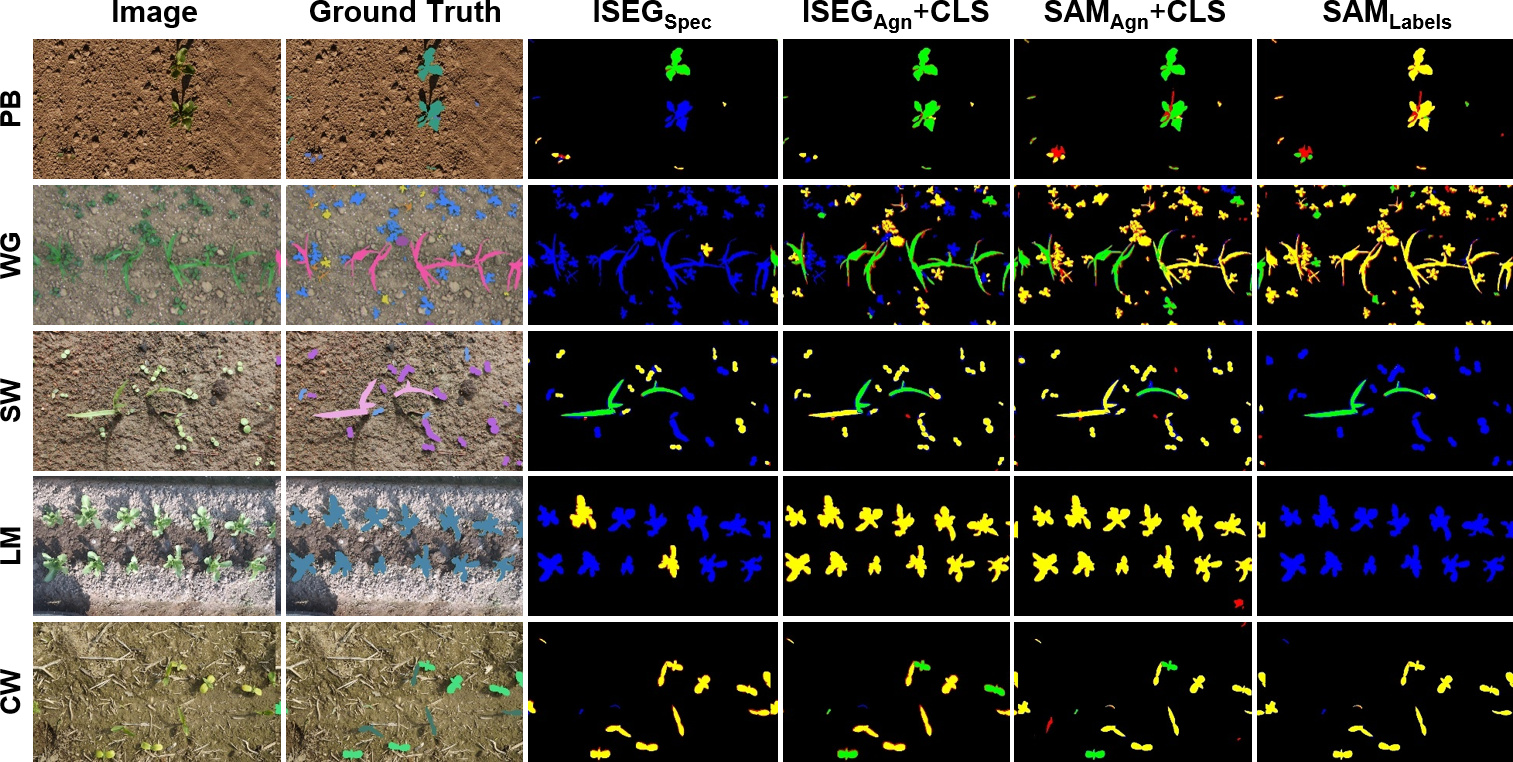}
    \caption{Representative unsupervised/weakly supervised upcycling results on images from \textbf{P}heno\textbf{B}ench \cite{WeMa24}, \textbf{W}eeds\textbf{G}alore \cite{CeKu25}, \textbf{S}orghum\textbf{W}eedSeg \cite{JuTh24}, \textbf{L}ettuce\textbf{M}OTS \cite{HuSu24} and \textbf{C}ropAnd\textbf{W}eed \cite{StTr23}. Left to right: Input images, manual annotations, results by method (green: correctly detected mask areas, yellow: correctly located areas assigned to wrong classes, red: incorrectly detected areas, blue: undetected instances).}
    \label{fig:qualitative_comparison}
\end{figure}

However, the preferred approach strongly depends on the characteristics of the data to be upcycled. Naturally, classification models are limited by the classes they are trained on. While most of our models perform well at classifying common plant species across datasets, such as sugar beets in \textit{PhenoBench}, the domain-specific classifier fails to detect lettuce plants in \textit{LettuceMOTS}, as they do not appear in the training data sourced from other datasets. While this extreme case is expected to become less common as the pool of integrated datasets expands, rare classes will always remain prone to under-representation. Using prior information in the form of label specifications (\textit{SAM\textsubscript{Labels}}) somewhat mitigates this issue for the case of \textit{LettuceMOTS}, but decreases performance for \textit{PhenoBench} due to SAM 3's less domain-specific training data. Performance on \textit{CropAndWeed} shows that high numbers of fine-grained classes pose a challenge for all strategies, but adding a separate classification stage improves precision compared to pure instance-segmentation, even if label information is available. For \textit{WeedsGalore}, classes available in the training data are assigned with high precision, but recall improves with fine-grained classification. Furthermore, as visible in \cref{tab:results_specific}, this is also the only case where a single-dataset baseline model outperforms our upcycling approaches despite their more extensive input data. However, this only applies for the model trained on \textit{CropAndWeed}, which provides highly fine-grained classes including those relevant for \textit{WeedsGalore}, while models trained on other datasets perform significantly worse than upcycling models. The influence of dataset characteristics on mask quality for all methods is further visualized by the qualitative results in \cref{fig:qualitative_comparison}.

\begin{table}[t]
    \caption{Instance-mask accuracy (average per-class \% mAP\textsuperscript{50-95}) and average inference \textbf{t}ime per image of upcycling methods on each target dataset (\textbf{P}heno\textbf{B}ench, \textbf{W}eeds\textbf{G}alore, \textbf{S}orghum\textbf{W}eedSeg, \textbf{L}ettuce\textbf{M}OTS, \textbf{C}ropAnd\textbf{W}eed). \textit{ISEG} and \textit{CLS} refer to models trained on aggregated \textit{ReLEAP} data for instance segmentation and classification, \textit{SAM} to SAM 3 prompted with priors in subscript. \textit{Spec} and \textit{Agn} denote multi-class and single-superclass models. The baseline represents average and best results of individual models trained on the remaining four datasets for each target.}
    \label{tab:results_specific}
    \centering
    \resizebox{1\linewidth}{!}{
    \begin{tabular}{@{}l@{\qquad}l@{\quad}r@{\hspace{0.5em}}r@{\hspace{0.5em}}r@{\hspace{0.5em}}r@{\hspace{0.5em}}r@{\hspace{0.5em}}r@{\quad}r@{}}
        \toprule
        \textbf{Method} & \textbf{Supervision} & \multicolumn{1}{l@{\quad}}{\textbf{PB}} & \textbf{WG} & \textbf{SW} & \textbf{LM} & \textbf{CW} & \textbf{Overall} & \multicolumn{1}{@{\quad}r@{}}{\textbf{t [ms]}} \\        
        \midrule
        \textit{Baseline\textsubscript{Overall}} & & 9.9 & 4.0 & 0.8 & 0.0 & 0.2 & 3.0 & 34.6 \\
        \textit{Baseline\textsubscript{Best}} & & 34.8 & \textbf{15.6} & 2.3 & 0.0 & 0.5 & \textbf{10.6} & \textbf{31.4} \\
        \textit{ISEG\textsubscript{Spec}} & Unsupervised & 31.9 & 10.6 & 2.7 & 0.0 & \textbf{0.9} & 9.2 & 33.9 \\
        \textit{SAM\textsubscript{Agn}}+\textit{CLS} & & 35.7 & 7.9 & 1.2 & 0.0 & 0.1 & 9.0 & 727.9 \\
        \textit{ISEG\textsubscript{Agn}}+\textit{CLS} & & \textbf{42.4} & 7.1 & \textbf{3.1} & 0.0 & 0.1 & 10.5 & 94.2 \\  
        \midrule
        \textit{SAM\textsubscript{Labels}} & Weakly supervised & 4.5 & 0.7 & 2.4 & 57.6 & 0.1 & 13.1 & 1,367.4 \\       
        \midrule
        \textit{SAM\textsubscript{Pt}} & Semi-supervised & 16.6 & 7.3 & 3.2 & 84.9 & 16.4 & 25.7 & 678.7 \\    
        \midrule
        \textit{ISEG\textsubscript{Agn}}+\textit{Box} & & 51.0 & 18.1 & 24.6 & 87.8 & 31.9 & 42.7 & \textbf{46.3} \\
        \textit{SAM\textsubscript{Agn}}+\textit{Box} & Semi-supervised & 43.9 & 25.1 & 18.9 & 86.8 & 40.8 & 43.1 & 667.9 \\
        \textit{SAM\textsubscript{Box}} &  & \textbf{67.3} & \textbf{40.6} & \textbf{32.6} & \textbf{89.4} & \textbf{49.9} & \textbf{55.9} & 686.9 \\
        \bottomrule
    \end{tabular}
    }
\end{table}

Overall, a domain-specific multi-class model achieves high precision on classes present in both training data and target images, but has no means of detecting anything outside this scope. Adding a separate classification stage considerably expands the pool of available classes, leading to higher recall at the same level of precision. Prompting SAM 3 with label specifications leads to slightly improved results compared to domain-specific classification, but requires prior information that might not always be available. Furthermore, as visible in \cref{tab:results_specific}, involving foundation models always leads to drastically higher runtime, while a combination of two domain-specific models provides a more lightweight solution.

\subsubsection{Semi-Supervised Approaches.}
When existing lower-level annotations are available for the target dataset, using them to prompt SAM 3 can be a worthwhile solution despite higher processing times, as visible in \cref{fig:methods_pr} and \cref{tab:results_specific}. For this scenario, the task becomes considerably easier, as label assignment is usually already included in the annotations of bounding boxes or point annotations defining instances. The former prove to be the most useful modality for prompting SAM 3, as the task of delineating object contours of individual plants within specified image regions can be automated with the highest accuracy by far. However, this result should be interpreted as an upper-performance reference, as \textit{SAM\textsubscript{Box}} assumes the availability of bounding-box annotations and therefore operates under substantially stronger supervision than the unsupervised and weakly supervised approaches. Point prompts, on the other hand, lead to inferior results to all box-based strategies, only marginally surpassing unsupervised approaches for most datasets.

Prompting SAM 3 only with generic superclasses or applying a domain-specific class-agnostic model to generate masks and then using existing bounding boxes for label assignment (\textit{SAM\textsubscript{Agn}}+\textit{Box}, \textit{ISEG\textsubscript{Agn}}+\textit{Box}) does not provide an advantage over direct prompting with the same boxes for most datasets. However, the prior information considerably improves accuracy compared to unsupervised classification. This is to be expected, since the agnostic methods already proved to handle the localization of plant instances reasonably well, as visible in \cref{tab:results_agnostic}, and are now combined with a relatively stable means of class assignment. For this scenario, domain-specific models are preferable to foundation models due to their similar performance combined with drastically reduced inference time.

\subsection{Few-Shot Domain Adaptation}
\label{sec:few_shot}
As an alternative to using a dedicated classification model, we evaluate how domain adaptation between datasets can be improved by few-shot learning based on our class-specific \textit{ISEG\textsubscript{Spec}} instance-segmentation models. We select the \textit{LettuceMOTS} and \textit{CropAndWeed} datasets for this purpose, for which the models showed lowest performance in the previous experiments, albeit presumably due to very different reasons. While \textit{LettuceMOTS} data poses a relatively simple target for detection with a single class and uniform image composition, the approach failed simply because the relevant class was not in the aggregated training data. \textit{CropAndWeed}, on the other hand, is an extremely challenging target due to its high number of classes depicted under variable environmental conditions.

Since we aim to emulate an upcycling scenario without prior information of the new image data, we cannot sample classes evenly but only use different numbers of random images, which would be manually annotated in a corresponding use case. For each experiment, we sample a support set of $K$ images, then fine-tune our original models of each target dataset for 50 epochs with default hyper-parameters. The goal is to measure how performance improves as a function of $K$, as shown in \cref{tab:results_fine_tuning}.

\begin{table}
    \caption{Accuracy (average per-class \% mAP\textsuperscript{50-95}) of few-shot domain-adaptation using class-specific models fine-tuned on different numbers of randomly selected image samples from two target datasets. Each value represents the mean of three experiment runs, along with standard deviation. Original results with no additional samples are included for reference, as the configuration slightly differs from previous evaluations.}
    \label{tab:results_fine_tuning}
    \centering
    \begin{tabular}{@{}l@{\qquad}r@{\quad}r@{\quad}r@{\quad}r@{\quad}r@{}}
        \toprule
        \multicolumn{1}{@{}l@{\qquad}}{\textbf{Target dataset}} 
            & \multicolumn{1}{c@{\quad}}{\textbf{0}} 
            & \multicolumn{1}{c@{\quad}}{\textbf{25}} 
            & \multicolumn{1}{c@{\quad}}{\textbf{50}} 
            & \multicolumn{1}{c@{\quad}}{\textbf{100}} 
            & \multicolumn{1}{c@{}}{\textbf{200}} \\           
        \midrule
        \textit{CropAndWeed} \cite{StTr23} & 1.5 & 1.5\textsuperscript{ $\pm~0.1$} & 2.6\textsuperscript{ $\pm~0.2$} & 2.7\textsuperscript{ $\pm~0.0$} & 3.2\textsuperscript{ $\pm~0.1$} \\        
        \textit{LettuceMOTS} \cite{HuSu24} & 0.0 & 48.0\textsuperscript{ $\pm~0.9$} & 90.4\textsuperscript{ $\pm~1.9$} & 91.8\textsuperscript{ $\pm~0.1$} & 92.2\textsuperscript{ $\pm~0.1$} \\
        \bottomrule
    \end{tabular}
\end{table}

As expected, the benefits of providing annotations for a small set of images from the data to be upcycled are more pronounced for the structurally simpler \textit{LettuceMOTS} dataset. A previously unrepresented class can readily be added with relatively few training samples. For \textit{CropAndWeed}, on the other hand, performance increases are not as significant, even for the highest number of samples. This is to be expected, as a random subset of this size will usually not cover all classes in the dataset missing from the original training setup. These results quantify how the benefits gained from few-shot domain adaptation differ depending on the target dataset and its class distribution and thereby provide a further hint for selecting suitable upcycling strategies for different scenarios.
\section{Conclusion}
\label{sec:conclusion}
We introduced \textit{ReLEAP}, a framework for unifying heterogeneous vision datasets and extending their utility through annotation recycling. By abstracting diverse data sources, harmonizing label spaces through a shared taxonomy and integrating 15 agricultural datasets, it provides a coherent foundation for combining data across tasks and domains with minimal manual effort. Our recycling mechanisms further demonstrate how seamlessly integrating deterministic task conversions can substantially expand the availability of training data.

Building on this unified data foundation, we investigated multiple strategies for upcycling annotations to higher-level tasks and automatically labeling previously unseen image data under varying degrees of supervision. Our results show that aggregating data from multiple sources consistently improves class-agnostic mask quality, while fine-grained class assignment benefits further from separating mask generation from classification to exploit the much larger pool of recycled classification data. Both outperform single-dataset training on average. Given the same prior information, lightweight domain-specific models perform comparably to much larger foundation models while their inference time is lower by roughly an order of magnitude. Furthermore, we quantified how prior information, model choice, task conversions and dataset characteristics influence annotation quality, providing practical guidance on selecting suitable upcycling strategies for different application scenarios. Together, these findings demonstrate the potential of reusing and extending existing data resources instead of creating new annotations from scratch, supporting more accessible, sustainable and resource-efficient use of vision data, even in settings with limited annotation capacity or computational resources.

We envision \textit{ReLEAP} as a community-driven resource that continues to grow with new datasets, taxonomies and upcycling methods. By fostering unified data access and reusable annotation pipelines, we hope to further improve the efficiency, accessibility and reproducibility of research across precision agriculture and potentially other heterogeneous vision domains.

\paragraph{Acknowledgments}
This research was funded by the Austrian Research Promotion Agency (FFG) under the BRIDGE-Project PADUA (FO999915101). We thank Farm-ING and WALDLAND for their continued collaboration and valuable domain expertise, which helped inform the practical context of this work.

\bibliographystyle{splncs04}
\bibliography{references}

\newpage

\begin{center}
    \Large
    \textbf{ReLEAP: Upcycling Vision Datasets through Cross-Task Unification and Semi-Supervised Annotation (Supplementary Material)} \\
    \vspace{2em}
\end{center}

\appendix
This supplementary document complements the main paper with additional details regarding dataset statistics, upcycling methods and experiment results. \cref{sec:data_unification_supp} provides statistics on surveyed and integrated datasets for \textit{ReLEAP}, as well as a comparison to other data-harmonization frameworks. \cref{sec:upcycling_methods_supp} contains implementation details of our upcycling methods and label mappings for model training, while \cref{sec:expirental_results_supp} shows detailed quantitative and qualitative results of the corresponding cross-dataset experiments.

\section{Data Unification and Recycling}
\label{sec:data_unification_supp}
For a more thorough understanding of dataset selection and integration considerations, this section provides detailed statistics of surveyed and integrated datasets. Furthermore, \textit{ReLEAP}'s features are positioned against other data-harmonization frameworks.

\subsection{Dataset Survey and Selection}
\begin{figure}
    \centering
    \includegraphics[width=1\columnwidth]{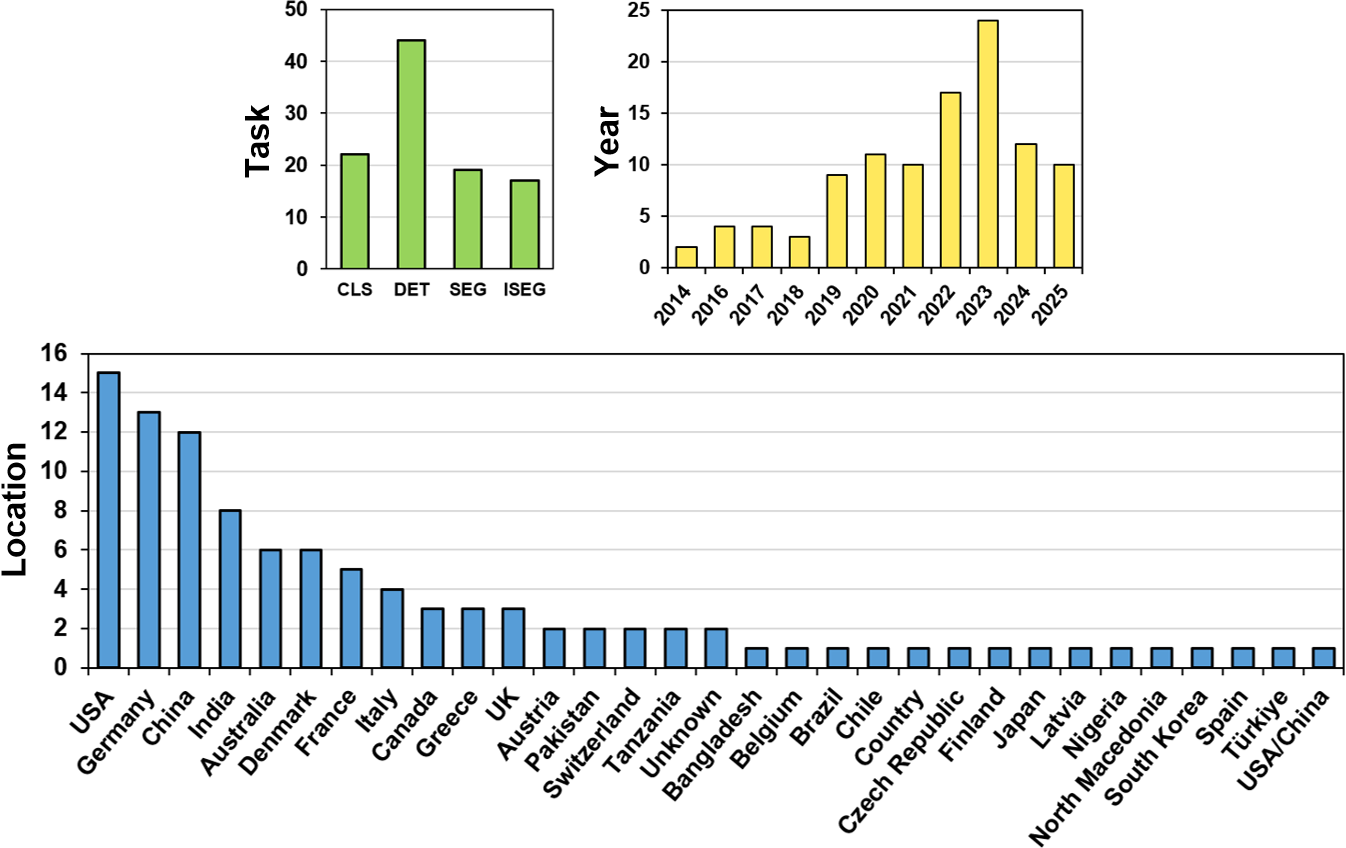}
    \caption{Distribution of surveyed datasets regarding their highest-level explicitly annotated learning tasks (CLS: classification, DET: object detection, SEG: semantic segmentation, ISEG: instance segmentation), as well as years and locations of recording.}
    \label{fig:datasets_superstats}
\end{figure}

The current selection of dataset interfaces in \textit{ReLEAP} is the result of an extensive survey of publicly available sources providing images and annotations of plants in early growth stages. They were thoroughly analyzed regarding their recording and annotation characteristics to form a balanced initial selection for evaluating our framework and its potential as a basis for data recycling and upcycling tasks. \cref{fig:datasets_superstats} gives an overview of the surveyed datasets' distributions across learning tasks, recording times and locations. Details for each dataset are presented in \cref{tab:cls_datasets,tab:det_datasets,tab:seg_datasets,tab:iseg_datasets}, categorized by the highest-level learning task for which explicit annotations are provided by the original source. Based on this survey, we selected the 15 datasets listed in \cref{tab:stats_selected_datasets} of the main paper for initial integration and experiments.

\begin{table}
    \caption{List of surveyed datasets annotated for classification as their highest-level task, which were considered for integration in \textit{ReLEAP}, along with recording years and locations, numbers of instances and species (Spec) as well as public availability (PA). Datasets selected for integration are marked in bold.}
    \label{tab:cls_datasets}
    \centering
    \setlength{\tabcolsep}{4pt}
    \begin{tabular}{@{}lclrrc@{}}
        \toprule    
        \textbf{Dataset} & \textbf{Year} & \textbf{Location} & \textbf{Instances} & \textbf{Spec} & \textbf{PA} \\
        \midrule
        AIWeeds \cite{DuZh22} & 2021 & USA/China & 10,000 & 15 & \xmark \\
        BeanDiseases \cite{LaMd24} & 2023 & Tanzania & 59,069 & 1 & \checkmark \\
        CornWeedCls \cite{Zh19} & 2019 & China & 5,998 & 5 & \checkmark \\
        CottonLeafDisease \cite{BiNi24} & 2024 & Bangladesh & 2,137 & 1 & \checkmark \\
        \textbf{CottonWeedID15} \cite{ChLu22} & \textbf{2022} & \textbf{USA} & \textbf{5,187} & \textbf{15} & \checkmark \\
        CWD30 \cite{IlAr23} & 2023 & South Korea & 219,778 & 30 & \checkmark \\
        DeepWeeds \cite{OlKo19} & 2019 & Australia & 17,509 & 8 & \checkmark \\
        DND-SB \cite{YiKr20} & 2020 & Germany & 5,647 & 1 & \checkmark \\
        \textbf{EarlyCropWeed} \cite{EsBo20} & \textbf{2020} & \textbf{Greece} & \textbf{508} & \textbf{4} & \checkmark \\
        LettuceWeed \cite{Zh19} & 2019 & China & 795 & 2 & \checkmark \\
        PlantLabSubset \cite{BeLi21} & 2021 & Canada & 14,000 & 14 & \checkmark \\
        \textbf{PlantSeedlings} \cite{GiMo17} & \textbf{2017} & \textbf{Denmark} & \textbf{5,539} & \textbf{12} & \checkmark \\
        PlantVillage \cite{MoHu16} & 2016 & Switzerland & 54,305 & 14 & \checkmark \\
        PotatoBlt \cite{LaMd25} & 2023 & Tanzania & 58,709 & 1 & \checkmark \\
        RadishWeed \cite{Zh19} & 2019 & China & 401 & 2 & \checkmark \\
        SorghumWeedCls \cite{JuTh24} & 2024 & India & 4,312 & 3 & \checkmark \\
        SpotWeeds \cite{RaZh22} & 2022 & USA & 30,815 & 6 & \xmark \\
        Weed25 \cite{WaTa22} & 2022 & China & 14,035 & 25 & \checkmark \\
        WeedGrowthState \cite{TeDy18} & 2018 & Denmark & 9,372 & 1 & \checkmark \\
        WeedID \cite{PeRe20} & 2020 & Germany & 93,130 & 12 & \xmark \\
        WeedNet \cite{ShAy25} & 2025 & USA & 150,154 & 1,593 & \xmark \\
        WeedSoybean \cite{FeFr17} & 2017 & Brazil & 15,336 & 3 & \checkmark \\
        \bottomrule
    \end{tabular}
    \centering
\end{table}

\begin{table}
    \caption{List of surveyed datasets annotated for object detection as their highest-level task, which were considered for integration in \textit{ReLEAP}, along with recording years and locations, numbers of images and species (Spec), integration of leaf-segmentation (L) or stem-point (S) annotations as well as public availability (PA). Datasets selected for integration are marked in bold.}
    \label{tab:det_datasets}
    \centering
    \resizebox{0.9\linewidth}{!}{
        \setlength{\tabcolsep}{4pt}
        \begin{tabular}{@{}lclrrcc@{}}
            \toprule
            \textbf{Dataset} & \textbf{Year} & \textbf{Location} & \textbf{Images} & \textbf{Spec} & \textbf{L/S} & \textbf{PA} \\
            \midrule            
            BelgiumBeet \cite{GaFr20} & 2020 & Belgium & 652 & 2 & & \xmark \\
            BroadleafWeeds \cite{Co21d} & 2022 & Australia & 78 & 1 & & \checkmark \\
            CanolaSeedling \cite{AnKr24} & 2024 & Canada & 431 & 1 & & \checkmark \\
            Chicory \cite{GaR23} & 2023 & Italy & 3,373 & 1 & & \checkmark \\
            Chilli \cite{RfCh23} & 2023 & Unknown & 707 & 2 & & \checkmark \\
            CobbityWheat \cite{Co21a} & 2021 & Australia & 140 & 3 & & \checkmark \\
            CornWeed \cite{Zh19} & 2023 & Germany & 3,574 & 2 & & \checkmark \\
            CottonWeedDet3 \cite{RaLu23} & 2022 & USA & 848 & 3 & & \checkmark \\
            CottonWeedDet12 \cite{DaCh23} & 2023 & USA & 5,648 & 12 & & \checkmark \\
            DeepSeedling \cite{JiLi19} & 2019 & USA & 5,741 & 2 & & \checkmark \\
            Eden \cite{MyMa22} & 2022 & Greece & 6,428 & 24 & & \checkmark \\
            FourWeed \cite{AgAh22} & 2022 & India & 618 & 4 & & \checkmark \\
            ImageWeeds \cite{RaMa23} & 2023 & USA & 3,975 & 6 & & \checkmark \\
            LettuceMOT \cite{HuWa22} & 2022 & China & 4,879 & 1 & & \checkmark \\
            LFSD \cite{WaSu23} & 2023 & China & 13,409 & 1 & & \xmark \\
            \textbf{LincolnBeet} \cite{SaDa22} & \textbf{2021} & \textbf{USA} & \textbf{4,402} & \textbf{2} & & \checkmark \\
            MaizeSet \cite{Al22} & 2023 & Pakistan & 3,330 & 1 & & \checkmark \\
            Maize-Weed \cite{OlSa23} & 2023 & Nigeria & 500 & 2 & & \checkmark \\
            MMDWWF \cite{XuJi22} & 2022 & China & 692 & 2 & & \xmark \\
            OpenPlantPhenotyping \cite{LeSi20} & 2020 & Denmark & 7,590 & 47 & & \checkmark \\
            PAGS8 \cite{Co21e} & 2023 & Australia & 614 & 1 & & \checkmark \\
            PaimioSB2023 \cite{PaKa24} & 2024 & Finland & 2,272 & 2 & & \checkmark \\
            \textbf{PakistaniWeed} \cite{KaPa22} & \textbf{2022} & \textbf{Pakistan} & \textbf{1,500} & \textbf{7} & & \checkmark \\
            PlantDoc \cite{SiJa20} & 2020 & India & 2,598 & 13 & & \checkmark \\
            RoboWeedMap \cite{TeJo22} & 2022 & Denmark & 1,147 & 2 & & \checkmark \\
            SesameWeeds \cite{Da20} & 2019 & India & 1,300 & 2 & & \checkmark \\
            SPAGRI-AI \cite{JoMu24} & 2024 & Czech Republic & 2,014 & 12 & & \checkmark \\
            \textbf{SugarbeetsAndWeeds} \cite{Wa21} & \textbf{2020} & \textbf{Unknown} & \textbf{120} & \textbf{2} & & \checkmark \\
            ThreeSeasonWeedDet10 \cite{DeLu24} & 2025 & USA & 1,784 & 10 & & \checkmark \\
            ThreeWeeds \cite{DeMi25} & 2025 & China & 1,982 & 3 & & \xmark \\
            TobSet \cite{AlAl22} & 2022 & Türkiye & 7,000 & 1 & & \checkmark \\
            TwoSeasonWeedDet8 \cite{DeLu24} & 2024 & USA & 6,664 & 21 & & \checkmark \\
            UAVMultiCrop \cite{DaDa21} & 2023 & France & 181 & 3 & & \checkmark \\
            \textbf{VegetableCrops} \cite{LaKe22} & \textbf{2022} & \textbf{France} & \textbf{2,610} & \textbf{3} & \textbf{S} & \checkmark \\
            WeedCropImages \cite{Sh20} & 2020 & Latvia & 1,118 & 14 & & \checkmark \\
            \textbf{WeedCrop }\cite{SuJa20} & \textbf{2025} & \textbf{USA} & \textbf{1,120} & \textbf{13} & & \checkmark \\
            Weed-yolov5 \cite{Is22} & 2022 & Unknown & 8,462 & 3 & & \checkmark \\
            WeedDetectionv2 \cite{Ta20} & 2021 & India & 210 & 8 & & \checkmark \\
            WeedGrowthStage \cite{IsSa25} & 2025 & USA & 203,567 & 16 & L & \checkmark \\
            WeedMaize \cite{CoTo21} & 2021 & Spain & 312 & 4 & & \checkmark \\
            WeedStemDetection \cite{LiLi25} & 2025 & China & 1,556 & 4 & & \checkmark \\
            WeedVision \cite{IsSa24} & 2025 & USA & 203,567 & 16 & & \checkmark \\
            WildCarrotInCanola \cite{Co21c} & 2021 & Chile & 52 & 1 & & \checkmark \\
            WildRadishInWeed \cite{Co21b} & 2021 & Australia & 41 & 1 & & \checkmark \\
            \bottomrule
        \end{tabular}
    }
\end{table}

\begin{table}
    \caption{List of surveyed datasets annotated for semantic segmentation as their highest-level task, which were considered for integration in \textit{ReLEAP}, along with recording years and locations, numbers of images and species (Spec), integration of leaf-segmentation (L) or stem-point (S) annotations as well as public availability (PA). Datasets selected for integration are marked in bold.}
    \label{tab:seg_datasets}
    \centering
    \setlength{\tabcolsep}{4pt}
    \begin{tabular}{@{}lclrrcc@{}}
        \toprule
        \textbf{Dataset} & \textbf{Year} & \textbf{Location} & \textbf{Images} & \textbf{Spec} & \textbf{L/S} & \textbf{PA} \\
        \midrule
        AgricultureVision \cite{ChXu20} & 2020 & USA & 94,986 & 2 & & \checkmark \\
        Airphen \cite{Va21} & 2020 & France & 275 & 2 & L & \checkmark \\
        Carrots2017 \cite{BoAp20} & 2019 & UK & 20 & 2 & & \checkmark \\
        Carrot-Weed \cite{LaZd17} & 2017 & North Macedonia & 39 & 2 & & \checkmark \\
        CoFly-WeedDB \cite{KrRa22} & 2022 & Greece & 201 & 3 & & \checkmark \\
        CWFID \cite{HaOs15} & 2014 & Germany & 60 & 2 & & \checkmark \\
        GrassClover \cite{SkDy19} & 2019 & Denmark & 435 & 4 & & \checkmark \\
        IndianPotatoWeeds \cite{GoNa23} & 2023 & India & 270 & 2 & & \checkmark \\
        Nassar \cite{WaHa23} & 2023 & Canada & 2,218 & 2 & & \checkmark \\
        Onions2017 \cite{BoAp20} & 2019 & UK & 20 & 2 & & \checkmark \\
        PWFID \cite{RvKr21} & 2021 & India & 150 & 2 & & \checkmark \\
        Redlands \cite{HaDa18} & 2018 & Australia & 1,907 & 4 & & \xmark \\
        ROSE \cite{BeMa23} & 2023 & France & 1,000 & 6 & & \checkmark \\
        SegPPD101 \cite{Ge25} & 2025 & China & 2,263 & 27 & & \checkmark \\
        \textbf{Sugarbeets2016} \cite{ChLo17} & \textbf{2016} & \textbf{Germany} & \textbf{11,539} & \textbf{2} & & \checkmark \\
        \textbf{Sunflower} \cite{PoIm24} & \textbf{2016} & \textbf{Italy} & \textbf{500} & \textbf{2} & & \checkmark \\
        UNL-MED \cite{DaDa23} & 2023 & USA & 130 & 1 & & \xmark \\
        VegAnn \cite{MaIr23} & 2023 & France & 3,712 & 1 & & \checkmark \\
        We3ds \cite{KiBa23} & 2023 & Austria & 2,568 & 18 & & \checkmark \\
        WeedNet \cite{SaCh17} & 2018 & Switzerland & 375 & 2 & & \checkmark \\
        \bottomrule
    \end{tabular}
\end{table}

\begin{table}
    \caption{List of surveyed datasets annotated for instance segmentation as their highest-level task, which were considered for integration in \textit{ReLEAP}, along with recording years and locations, numbers of images and species (Spec), integration of leaf-segmentation (L) or stem-point (S) annotations as well as public availability (PA). Datasets selected for integration are marked in bold.}
    \label{tab:iseg_datasets}
    \centering
    \setlength{\tabcolsep}{4pt}
    \begin{tabular}{@{}lclrrcc@{}}
        \toprule
        \textbf{Dataset} & \textbf{Year} & \textbf{Location} & \textbf{Images} & \textbf{Spec} & \textbf{L/S} & \textbf{PA} \\
        \midrule            
        Aberystwyth \cite{BeJo16} & 2016 & UK & 56 & 1 & L & \checkmark \\
        ACRECropWeed \cite{BeSp23} & 2023 & Italy & 1,000 & 6 & & \checkmark \\
        CN20 \cite{AhHa22} & 2022 & Germany & 283 & 9 & & \checkmark \\
        \textbf{CropAndWeed} \cite{StTr23} & \textbf{2023} & \textbf{Austria} & \textbf{8,034} & \textbf{74} & \textbf{S} & \checkmark \\
        Crops3D \cite{ZhZh24} & 2024 & China & 1,230 & 8 & & \checkmark \\
        GrowliFlower \cite{KiJu23} & 2023 & Germany & 2,198 & 1 & LS & \checkmark \\
        Komatsuna \cite{UcSa17} & 2017 & Japan & 1,200& 1 & L & \checkmark \\
        \textbf{LettuceMOTS} \cite{HuSu24} & \textbf{2023} & \textbf{China} & \textbf{1,308} & \textbf{1} & & \checkmark \\
        LSC \cite{ScMi14} & 2014 & Italy & 810 & 1 & L & \checkmark \\
        MFWD \cite{GeVa24} & 2024 & Germany & 907 & 31 & & \checkmark \\
        \textbf{PhenoBench} \cite{WeMa24} & \textbf{2023} & \textbf{Germany} & \textbf{2,179} & \textbf{2} & \textbf{L} & \checkmark \\
        RumexWeeds \cite{GuVa23} & 2022 & Denmark & 642 & 2 & & \checkmark \\
        SB20 \cite{AhHa21} & 2021 & Germany & 143 & 9 & & \checkmark \\
        SB21 \cite{AhHa24} & 2024 & Germany & 84 & 12 & & \xmark \\
        \textbf{SorghumWeedSeg} \cite{JuTh24} & \textbf{2024} & \textbf{India} & \textbf{252} & \textbf{3} & & \checkmark \\
        WeedAI \cite{HaZi24} & 2024 & Germany & 240 & 7 & & \checkmark \\
        \textbf{WeedsGalore} \cite{CeKu25} & \textbf{2025} & \textbf{Germany} & \textbf{156} & \textbf{5} & & \checkmark \\
        \bottomrule
    \end{tabular}
\end{table}

\clearpage
\subsection{Framework Comparison}

\cref{tab:framework_comparison} positions \textit{ReLEAP} relative to representative dataset-management and label-harmonization approaches. While existing tools support individual aspects such as dataset curation, visualization, format conversion or label-space alignment, \textit{ReLEAP} combines cross-dataset aggregation, taxonomy-based label harmonization and annotation recycling in a unified open-source workflow. Unlike \textit{Label remapping}, \textit{Domain-level taxonomy reuse} requires label equivalences to be specified once per application domain in the form of a shared structure onto which every dataset and every target label space is mapped, rather than as rules between a specific pair of label spaces. While the taxonomy itself is curated manually in all cases, the effort is incurred once per domain instead of per dataset. \textit{Task-driven annotation recycling} requires deterministic conversions between annotation types to be selected automatically from the requested target task during aggregation. \textit{End-to-end aggregation} is satisfied only where all capabilities above are provided within a single configurable pipeline.

\begin{table}
    \centering
    \caption{Qualitative comparison of \textit{ReLEAP} with related dataset-management and label-harmonization frameworks (Roboflow \cite{DwNe26}, FiftyOne \cite{Vo26}, Datumaro \cite{Op26} and MSeg \cite{LaLi20}). Marks denote whether a capability is available as a documented part of the respective workflow, not its maturity or scope. The label \textit{partial} indicates support for selected aspects, but not as a primary objective or through an automated end-to-end workflow. \textit{ReLEAP} combines cross-dataset integration, taxonomy-based label harmonization and task-driven annotation recycling in a unified open-source framework.}
    \label{tab:framework_comparison}
    \resizebox{\linewidth}{!}{
        \setlength{\tabcolsep}{3pt}
        \begin{tabular}{lccccc} 
            \toprule 
            \textbf{Feature} & \textbf{\textit{ReLEAP}} & \textbf{Roboflow} & \textbf{FiftyOne} & \textbf{Datumaro} & \textbf{MSeg} \\
            \midrule
            Open-source core functionality   & \checkmark & partial & \checkmark  & \checkmark  & \checkmark  \\
            Dataset management               & \checkmark & \checkmark  & \checkmark  & \checkmark  & partial \\
            Dataset visualization            & \checkmark & \checkmark  & \checkmark  & partial & --      \\
            \midrule
            Annotation format conversion     & \checkmark & \checkmark  & \checkmark  & \checkmark  & --      \\
            Label remapping                  & \checkmark & \checkmark  & \checkmark  & \checkmark  & \checkmark  \\
            Domain-level taxonomy reuse      & \checkmark & --      & --      & --      & \checkmark  \\
            Hierarchical taxonomy support    & \checkmark & --      & --      & partial & --      \\
            Task-driven annotation recycling & \checkmark & --      & partial & partial & --      \\
            \midrule
            \textbf{End-to-end aggregation}  & \textbf{\checkmark} & partial & partial & partial & partial \\
            \bottomrule
        \end{tabular}
    }
\end{table}

\clearpage
\section{Upcycling Methods}
\label{sec:upcycling_methods_supp}
This section provides additional details on our upcycling methods and their exact configuration, as well as label mappings used for training models.

To facilitate reproducibility of experimental results, the following list provides a comprehensive description of each evaluated upcycling method, along with exact configurations.

\paragraph{\textbf{Unsupervised Methods.}} This category contains all upcycling approaches not requiring any prior information in the form of existing annotations. They operate solely on the image data to be upcycled, either using models trained on the existing data of all other available sources, or by using foundation models.

\begin{itemize}
    \setlength\itemsep{0.55em}
    \item \textbf{\textit{ISEG\textsubscript{Spec}}}: This approach is the most straight-forward, as it consists of simply training a model on the available instance-segmentation data of all integrated datasets, except for the target dataset used in the respective experimental round. Data preparation and model training is conducted according to the description in \cref{sec:evaluation_protocol} of the main paper and inference uses a confidence threshold of 0.4, which showed a reasonable tradeoff between recall and precision for the purpose of semi-automatic annotation.
    
    \item \textbf{\textit{ISEG\textsubscript{Agn}}}+\textit{\textbf{CLS}}: In this case, two separate models are applied for each experiment round. The first is trained in the same way as for \textit{ISEG\textsubscript{Spec}}, but with all plant species mapped to a single superclass. This results in a slightly higher number of available training samples, as they are not filtered to exclude underrepresented classes. The class-agnostic instance masks retrieved by this model on the target dataset are used to crop patches from the input images using their bounding boxes, which form the inputs for class assignment by a separate classification model. As described in \cref{sec:evaluation_protocol} of the paper, the classification model for each round can be trained on a significantly higher number of input datasets, but also excludes all samples recycled from the respective target dataset. The confidence thresholds for inference are set to 0.4 for instance segmentation and 0.1 for classification with unclassified instances removed from the results. As localization by the agnostic models is relatively robust, the classifier can be set to this low threshold for filtering.
    
    \item \textbf{\textit{SAM\textsubscript{Agn}}}+\textit{\textbf{CLS}}: This setup uses the same classification model as above, but with agnostic masks generated by prompting SAM 3 \cite{CaGu25} with the generic superclass names \textit{plant}, \textit{vegetation}, \textit{crop} and \textit{weed} and applying non-maxima suppression on the combined results based on an intersection-over-union (IoU) threshold of 0.7. For interfacing with SAM 3, we use the API provided by Ultralytics \cite{JoQi24} configured to use half precision and a confidence threshold of 0.25.
\end{itemize}

\paragraph{\textbf{Weakly Supervised Methods.}} Methods in this category do not require existing annotations on the level of individual images, but only meta-information about the entire dataset. In this case, we evaluate a single strategy based on the prior information of label names expected to occur in the target images.

\begin{itemize}
    \setlength\itemsep{0.55em}
    \item \textbf{\textit{SAM\textsubscript{Labels}}}: This approach uses the same configuration of SAM 3 as for the \textit{SAM\textsubscript{Agn}} model, but with different text prompts. Instead of generic superclass names, we separately prompt the model with each class name specified by the original source of the respective target dataset, excluding generic background labels, such as \textit{soil} or \textit{background}, and use the same non-maxima-suppression threshold as for \textit{SAM\textsubscript{Agn}} across all labels. This is equivalent to a use case for upcycling completely unannotated image data with some prior knowledge regarding the recorded crop and weed species.
\end{itemize}

\paragraph{\textbf{Semi-Supervised Methods.}} This category summarizes upcycling approaches relying on prior annotations for lower-level tasks assigned to each image in the form of bounding boxes or stem points, which make localization and label assignment significantly less challenging.

\begin{itemize}
    \setlength\itemsep{0.55em}
    \item \textbf{\textit{ISEG\textsubscript{Agn}}}+\textit{\textbf{Box}}: This approach applies the same \textit{ISEG\textsubscript{Agn}} models as the corresponding unsupervised method with the same configuration, but instead of using their class-agnostic results as inputs for a classification model, labels can be assigned based on bounding-box annotations. For this purpose, we attempt to assign one of the existing boxes to the bounding box derived from each detected mask by IoU-based matching with a threshold of 0.1, allowing the assignment of the same annotation box to multiple detected masks. Labels are then directly assigned from the best-matching boxes of each mask, while unmatched masks are discarded.
    
    \item \textbf{\textit{SAM\textsubscript{Agn}}}+\textit{\textbf{Box}}: In this case, the \textit{SAM\textsubscript{Agn}} model from the corresponding unsupervised method is used in combination with the same label-assignment algorithm as for \textit{ISEG\textsubscript{Agn}}+\textit{Box}.
    
    \item \textbf{\textit{SAM\textsubscript{Box}}}: Alternatively to the above label assignment, existing bounding boxes can be directly used as prompts for SAM 3. The masks generated within each box are then assigned the original boxes' annotation labels.
    
    \item \textbf{\textit{SAM\textsubscript{Pt}}}: This approach is suitable for image data with existing annotations of stem-points for each plant. Since these annotations are not available for all our target datasets, we simulate them by geometrically deriving a point within the ground-truth mask of each plant instance as prompting input for SAM 3 and assigning the corresponding labels to the results.
\end{itemize}

\subsection{Label Mapping}
Prior to training multi-class models for instance segmentation (\textit{ISEG\textsubscript{Spec}}) and classification (\textit{CLS}), the classes and instances of datasets integrated in \textit{ReLEAP} are mapped and filtered according to the description in \cref{sec:evaluation_protocol} of the main paper. This results in the label definitions listed in \cref{tab:classes_iseg} for instance segmentation and \cref{tab:classes_cls_1,tab:classes_cls_2} for classification. \cref{tab:iseg_stats,tab:cls_stats} give a detailed view of label and instance counts for each target dataset after mapping and filtering.

\begin{table}
    \caption{Class definition for \textbf{multi-class instance-segmentation} models, along with botanical names and EPPO \cite{Ep26} codes from our hierarchical taxonomy, total numbers of instances across the five selected datasets and their presence in each of them (\textit{PhenoBench} (PB), \textit{WeedsGalore} (WG), \textit{SorghumWeedSeg} (SW), \textit{LettuceMOTS} (LM) and \textit{CropAndWeed} (CW)).}
    \label{tab:classes_iseg}
    \centering
    \resizebox{0.75\linewidth}{!}{
        \setlength{\tabcolsep}{4pt}
        \begin{tabular}{@{}llccr@{}}
            \toprule
            \textbf{Class name} & \textbf{Botanical name} & \textbf{EPPO} & \textbf{Datasets} & \textbf{Instances} \\
            \midrule
            AnnualMeadowGrass & Poa annua & POAAN & CW & 359 \\
            AnnualMercury & \textit{Mercurialis annua} & MERAN & CW & 894 \\
            BarnyardGrass & \textit{Echinochloa} & 1ECHG & WG, CW & 3,429 \\
            BlackBindweed & \textit{Fallopia convolvulus} & POLCO & CW & 200 \\
            BroadBean & \textit{Vicia faba} & VICFX & CW & 917 \\
            CharlockMustard & \textit{Sinapis arvensis} & SINAR & CW & 330 \\
            CommonBuckwheat & \textit{Fagopyrum esculentum} & FAGES & CW & 187 \\
            CommonChickweed & \textit{Stellaria media} & STEME & CW & 174 \\
            CommonCorncockle & \textit{Agrostemma githago} & AGOGI & CW & 1,579 \\
            CommonFumitory & \textit{Fumaria officinalis} & FUMOF & CW & 212 \\
            CommonKnotgrass & \textit{Polygonum aviculare} & POLAV & CW & 366 \\
            CommonPoppy & \textit{Papaver rhoeas} & PAPRH & CW & 505 \\
            CommonSunflower & \textit{Helianthus annuus} & HELAN & CW & 1,793 \\
            CommonWildOat & \textit{Avena fatua} & AVEFA & CW & 392 \\
            CopseBindweed & \textit{Fallopia dumetorum} & BIKDU & CW & 1,519 \\
            Cornflower & \textit{Centaurea cyanus} & CENCY & CW & 1,794 \\
            CornSpurry & \textit{Spergula arvensis} & SPRAR & CW & 758 \\
            CreepingThistle & \textit{Cirsium arvense} & CIRAR & CW & 2,522 \\
            Dicots & \textit{Dicotyledons} & 3DICOT & SW & 2,698 \\
            FieldBindweed & \textit{Convolvulus arvensis} & CONAR & CW & 472 \\
            FieldPennycress & \textit{Thlaspi arvense} & THLAR & CW & 106 \\
            FieldSowthistle & \textit{Sonchus arvensis} & SONAR & CW & 2,074 \\
            Geraniums & \textit{Geranium} & 1GERG & CW & 669 \\
            Grasses & Cyperaceae & 1GRAF & SW, CW & 676 \\
            GreenFoxtail & \textit{Setaria viridis} & SETVI & CW & 5,801 \\
            HairyCrabgrass & \textit{Digitaria sanguinalis} & DIGSA & CW & 405 \\
            HedgeMustard & \textit{Sisymbrium officinale} & SSYOF & CW & 727 \\
            IvyLeavedSpeedwell & \textit{Veronica hederifolia} & VERHE & CW & 225 \\
            Lettuce & \textit{Lactuca sativa} & LACSA & LM & 17,588 \\
            Maize & \textit{Zea mays} & ZEAMX & WG, CW & 8,214 \\
            MapleLeafGoosefoot & \textit{Chenopodium hybridum} & CHEHY & CW & 261 \\
            MeadowGrasses & \textit{Poa} & 1POAG & CW & 2,703 \\
            MouseBarley & \textit{Hordeum murinum} & HORMU & CW & 446 \\
            Pea & \textit{Pisum sativum} & PIBSX & CW & 594 \\
            Peppermint & \textit{Mentha x piperita} & MENPI & CW & 337 \\
            Pineappleweed & \textit{Matricaria discoidea} & MATMT & CW & 458 \\
            Potato & \textit{Solanum tuberosum} & SOLTU & CW & 366 \\
            PotatoWeed & \textit{Galinsoga parviflora} & GASPA & WG, CW & 2,422 \\
            RedRootAmaranth & \textit{Amaranthus retroflexus} & AMARE & WG, CW & 6,959 \\
            Redshank & \textit{Persicaria maculosa} & POLPE & CW & 402 \\
            RibwortPlantain & \textit{Plantago lanceolata} & PLALA & CW & 736 \\
            RyeBrome & \textit{Bromus secalinus} & BROSE & CW & 564 \\
            SmallGeranium & \textit{Geranium pusillum} & GERPU & CW & 3,334 \\
            SoftBrome & \textit{Bromus hordeaceus} & BROMO & CW & 349 \\
            Sorghum & \textit{Sorghum} & 1SORG & SW & 1,582 \\
            Soybean & \textit{Glycine max} & GLXMA & CW & 3,944 \\
            Squash & \textit{Cucurbita} & 1CUUG & CW & 608 \\
            SugarBeet & \textit{Beta vulgaris var. altissima} & BEAVA & PB, CW & 27,242 \\
            Thornapple & \textit{Datura stramonium} & DATST & CW & 455 \\
            Vegetation & - & - & CW & 5,955 \\
            Weed & - & - & PB, WG & 12,926 \\
            WhiteGoosefoot & \textit{Chenopodium album} & CHEAL & CW & 2,440 \\
            WildChamomile & \textit{Matricaria chamomilla} & MATCH & CW & 2,057 \\
            WildPansy & \textit{Viola tricolor} & VIOTR & CW & 232 \\
            Wintercress & \textit{Barbarea vulgaris} & BARVU & CW & 355 \\
            \bottomrule
        \end{tabular}
    }
\end{table}

\begin{table}
    \caption{Class definition for \textbf{classification} models, along with botanical names and EPPO \cite{Ep26} codes from our hierarchical taxonomy and total numbers of instances across the thirteen classification datasets (Part 1).}
    \label{tab:classes_cls_1}
    \centering
    \resizebox{0.85\linewidth}{!}{
        \setlength{\tabcolsep}{4pt}
        \begin{tabular}{@{}llcr@{}}
            \toprule
            \textbf{Class name} & \textbf{Botanical name} & \textbf{EPPO} & \textbf{Instances} \\
            \midrule
            AngledLuffa & \textit{Luffa acutangula} & LUFAC & 2,768 \\
            AnnualMeadowGrass & \textit{Poa annua} & POAAN & 415 \\
            AnnualMercury & \textit{Mercurialis annua} & MERAN & 974 \\
            BarnyardGrass & \textit{Echinochloa} & 1ECHG & 2,262 \\
            BitterMelon & \textit{Momordica charantia} & MOMCH & 2,116 \\
            BlackBean & \textit{Phaseolus vulgaris ssp. nanus} & PHSVN & 1,970 \\
            BlackBindweed & \textit{Fallopia convolvulus} & POLCO & 193 \\
            BlackGrass & \textit{Alopecurus myosuroides} & ALOMY & 309 \\
            BlackNightshade & \textit{Solanum nigrum} & SOLNI & 137 \\
            BroadBean & \textit{Vicia faba} & VICFX & 936 \\
            CannabisIndica & \textit{Cannabis indica} & CNISI & 7,453 \\
            CharlockMustard & \textit{Sinapis arvensis} & SINAR & 817 \\
            ChineseSenna & \textit{Senna obtusifolia} & CASOB & 240 \\
            Cleavers & \textit{Galium aparine} & GALAP & 369 \\
            CommonBean & \textit{Phaseolus vulgaris} & PHSVX & 3,537 \\
            CommonBuckwheat & \textit{Fagopyrum esculentum} & FAGES & 196 \\
            CommonChickweed & \textit{Stellaria media} & STEME & 922 \\
            CommonCorncockle & \textit{Agrostemma githago} & AGOGI & 1,502 \\
            CommonCowslip & \textit{Primula veris} & PRIVE & 108 \\
            CommonFlax & \textit{Linum usitatissimum} & LIUUT & 1,745 \\
            CommonFumitory & \textit{Fumaria officinalis} & FUMOF & 221 \\
            CommonKnotgrass & \textit{Polygonum aviculare} & POLAV & 390 \\
            CommonPoppy & \textit{Papaver rhoeas} & PAPRH & 583 \\
            CommonPurslane & \textit{Portulaca oleracea} & POROL & 516 \\
            CommonRagweed & \textit{Ambrosia artemisiifolia} & AMBEL & 1,215 \\
            CommonSunflower & \textit{Helianthus annuus} & HELAN & 1,837 \\
            CommonWheat & \textit{Triticum aestivum} & TRZAX & 253 \\
            CommonWildOat & \textit{Avena fatua} & AVEFA & 365 \\
            CopseBindweed & \textit{Fallopia dumetorum} & BIKDU & 1,645 \\
            Cornflower & \textit{Centaurea cyanus} & CENCY & 1,691 \\
            CornSpurry & \textit{Spergula arvensis} & SPRAR & 873 \\
            Crabgrass & \textit{Digitaria} & 1DIGG & 111 \\
            CreepingThistle & \textit{Cirsium arvense} & CIRAR & 2,603 \\
            Eclipta & \textit{Eclipta} & 1ECLG & 254 \\
            FieldBindweed & \textit{Convolvulus arvensis} & CONAR & 510 \\
            FieldPennycress & \textit{Thlaspi arvense} & THLAR & 117 \\
            FieldSowthistle & \textit{Sonchus arvensis} & SONAR & 2,337 \\
            Geraniums & \textit{Geranium} & 1GERG & 648 \\
            GreenCarpetweed & \textit{Mollugo verticillata} & MOLVE & 763 \\
            GreenFoxtail & \textit{Setaria viridis} & SETVI & 8,682 \\
            HairyCrabgrass & \textit{Digitaria sanguinalis} & DIGSA & 436 \\
            HedgeMustard & \textit{Sisymbrium officinale} & SSYOF & 818 \\
            HerbParis & \textit{Paris quadrifolia} & PJRQU & 4,039 \\
            Horseweed & \textit{Erigeron canadensis} & ERICA & 5,255 \\
            \bottomrule
        \end{tabular}
    }
\end{table}

\begin{table}
    \caption{Class definition for \textbf{classification} models, along with botanical names and EPPO \cite{Ep26} codes from our hierarchical taxonomy and total numbers of instances across the thirteen classification datasets (Part 2).}
    \label{tab:classes_cls_2}
    \centering
    \resizebox{0.85\linewidth}{!}{
        \setlength{\tabcolsep}{4pt}
        \begin{tabular}{@{}llcr@{}}
            \toprule
            \textbf{Class name} & \textbf{Botanical name} & \textbf{EPPO} & \textbf{Instances} \\
            \midrule
            IndianGoosegrass & \textit{Eleusine indica} & ELEIN & 216 \\
            IvyLeavedSpeedwell & \textit{Veronica hederifolia} & VERHE & 219 \\
            Lentil & \textit{Lens culinaris} & LENCU & 1,706 \\
            Lettuce & \textit{Lactuca sativa} & LACSA & 17,151 \\
            LooseSilkyBent & \textit{Apera spica-venti} & APESV & 762 \\
            Maize & \textit{Zea mays} & ZEAMX & 11,710 \\
            MapleLeafGoosefoot & \textit{Chenopodium hybridum} & CHEHY & 313 \\
            MeadowGrasses & \textit{Poa} & 1POAG & 3,242 \\
            MorningGlorys & \textit{Ipomoea} & 1IPOG & 1,115 \\
            MouseBarley & \textit{Hordeum murinum} & HORMU & 426 \\
            Nutsedges & \textit{Cyperus} & 1CYPG & 273 \\            
            Okra & \textit{Abelmoschus esculentus} & ABMES & 15,509 \\
            PalmerAmaranth & \textit{Amaranthus palmeri} & AMAPA & 689 \\
            Pea & \textit{Pisum sativum} & PIBSX & 1,997 \\
            Peppermint & \textit{Mentha x piperita} & MENPI & 514 \\
            Pineappleweed & \textit{Matricaria discoidea} & MATMT & 563 \\
            Potato & \textit{Solanum tuberosum} & SOLTU & 365 \\
            PotatoWeed & \textit{Galinsoga parviflora} & GASPA & 2,580 \\
            PricklySida & \textit{Sida spinosa} & SIDSP & 129 \\
            Rapeseed & \textit{Brassica napus} & BRSNN & 1,804 \\
            RedRootAmaranth & \textit{Amaranthus retroflexus} & AMARE & 5,814 \\
            Redshank & \textit{Persicaria maculosa} & POLPE & 411 \\
            RibwortPlantain & \textit{Plantago lanceolata} & PLALA & 830 \\
            RoughfruitAmaranth & \textit{Amaranthus tuberculatus} & AMATU & 1,146 \\
            RoughMeadowGrass & \textit{Poa trivialis} & POATR & 378 \\
            RyeBrome & \textit{Bromus secalinus} & BROSE & 554 \\
            ScentlessFalseMayweed & \textit{Tripleurospermum inodorum} & MATIN & 607 \\
            ShepherdsPurse & \textit{Capsella bursa-pastoris} & CAPBP & 282 \\
            SmallGeranium & \textit{Geranium pusillum} & GERPU & 4,422 \\
            SoftBrome & \textit{Bromus hordeaceus} & BROMO & 348 \\
            Sorghum & \textit{Sorghum} & 1SORG & 1,657 \\
            Soybean & \textit{Glycine max} & GLXMA & 6,790 \\
            SpottedSpurge & \textit{Euphorbia maculata} & EPHMA & 234 \\
            Squash & \textit{Cucurbita} & 1CUUG & 605 \\
            SugarBeet & \textit{Beta vulgaris var. altissima} & BEAVA & 42,105 \\
            SummerCypress & \textit{Bassia scoparia} & KCHSC & 851 \\
            Thornapple & \textit{Datura stramonium} & DATST & 450 \\
            Tomato & \textit{Solanum lycopersicum} & LYPES & 201 \\
            Velvetleaf & \textit{Abutilon theophrasti} & ABUTH & 130 \\
            WhiteGoosefoot & \textit{Chenopodium album} & CHEAL & 3,496 \\
            WildChamomile & \textit{Matricaria chamomilla} & MATCH & 2,002 \\
            WildLeek & \textit{Allium ampeloprasum} & ALLAM & 3,068 \\
            WildPansy & \textit{Viola tricolor} & VIOTR & 252 \\
            Wintercress & \textit{Barbarea vulgaris} & BARVU & 362 \\
            \bottomrule
        \end{tabular}
    }
\end{table}

\begin{table}
    \caption{Numbers of mapped labels and included instances in target datasets for instance-segmentation experiments. Numbers of instances are provided for the class-agnostic \textbf{\textit{ISEG\textsubscript{Agn}}} variants filtered by size and for the multi-class \textbf{\textit{ISEG\textsubscript{Spec}}} variants additionally filtered by image and instance counts. Reductions in labels result both from mapping to superclasses and exclusion of underrepresented classes. Original values are provided for reference. As target labels overlap across datasets, their total is given as the number of distinct classes in the common label space (\cref{tab:classes_iseg}).}
    \label{tab:iseg_stats}
    \centering
    \begin{tabular}{@{}lrr@{\quad}rrr@{}}
        \toprule
            & \multicolumn{2}{c@{\quad}}{\textbf{Labels}} 
            & \multicolumn{3}{c}{\textbf{Instances}} \\
        \textbf{Dataset}
            & \textbf{Original} & \textbf{Target}
            & \textbf{Original} & \textbf{Agnostic} & \textbf{Specific} \\
        \midrule
        \textit{CropAndWeed} \cite{StTr23} & 75 & 51 & 109,794 & 74,002 & 70,113 \\
        \textit{LettuceMOTS} \cite{HuSu24} & 1 & 1 & 18,370 & 17,588 & 17,588 \\
        \textit{PhenoBench} \cite{WeMa24} & 3 & 2 & 37,426 & 32,766 & 32,766 \\
        \textit{SorghumWeedSeg} \cite{JuTh24} & 3 & 3 & 5,051 & 4,888 & 4,888 \\
        \textit{WeedsGalore} \cite{CeKu25} & 5 & 5 & 12,779 & 9,957 & 9,957 \\
        \midrule
         & \textbf{87} & \textbf{55} & \textbf{183,420} & \textbf{139,201} & \textbf{135,312} \\
        \bottomrule
    \end{tabular}
\end{table}

\begin{table}
    \caption{Numbers of mapped labels and included instances in training datasets for the \textbf{\textit{CLS}} classification models after filtering by instance size. Reductions in labels result both from mapping to superclasses and exclusion of underrepresented classes. Original values are provided for reference. As target labels overlap across datasets, their total is given as the number of distinct classes in the common label space (\cref{tab:classes_cls_1,tab:classes_cls_2}).} 
    \label{tab:cls_stats}
    \centering
    \begin{tabular}{@{}lrr@{\quad}rr@{}}
        \toprule
            & \multicolumn{2}{c@{\quad}}{\textbf{Labels}} 
            & \multicolumn{2}{c}{\textbf{Instances}} \\
        \textbf{Dataset}
            & \textbf{Original} & \textbf{Target}
            & \textbf{Original} & \textbf{Specific} \\
        \midrule
        \textit{CottonWeedID15} \cite{ChLu22} & 15 & 13 & 5,187 & 5,054 \\
        \textit{CropAndWeed} \cite{StTr23} & 75 & 56 & 109,794 & 70,811 \\
        \textit{EarlyCropWeed} \cite{EsBo20} & 4 & 3 & 508 & 454 \\
        \textit{LettuceMOTS} \cite{HuSu24} & 1 & 1 & 18,370 & 17,151 \\
        \textit{LincolnBeet} \cite{SaDa22} & 2 & 1 & 39,246 & 16,342 \\
        \textit{PakistaniWeed} \cite{KaPa22} & 7 & 6 & 42,840 & 36,733 \\
        \textit{PhenoBench} \cite{WeMa24} & 3 & 1 & 37,426 & 17,856 \\
        \textit{PlantSeedlings} \cite{GiMo17} & 12 & 12 & 5,539 & 5,539 \\
        \textit{SorghumWeedSeg} \cite{JuTh24} & 3 & 1 & 5,051 & 1,657 \\
        \textit{SugarbeetsAndWeeds} \cite{Wa21} & 2 & 1 & 993 & 454 \\
        \textit{VegetableCrops} \cite{LaKe22} & 3 & 3 & 9,003 & 9,001 \\
        \textit{WeedCrop} \cite{SuJa20} & 13 & 13 & 17,693 & 17,693 \\
        \textit{WeedsGalore} \cite{CeKu25}  & 5 & 4 & 12,779 & 4,686 \\
        \midrule
         & \textbf{145} & \textbf{88} & \textbf{304,429} & \textbf{203,431} \\
        \bottomrule
    \end{tabular}
\end{table}

\clearpage

\section{Experimental Results}
\label{sec:expirental_results_supp}
This section provides additional details and exact values for our cross-dataset method evaluation. 

\subsubsection{Method Comparison.}
\begin{figure}
    \centering
    \includegraphics[width=1\columnwidth]{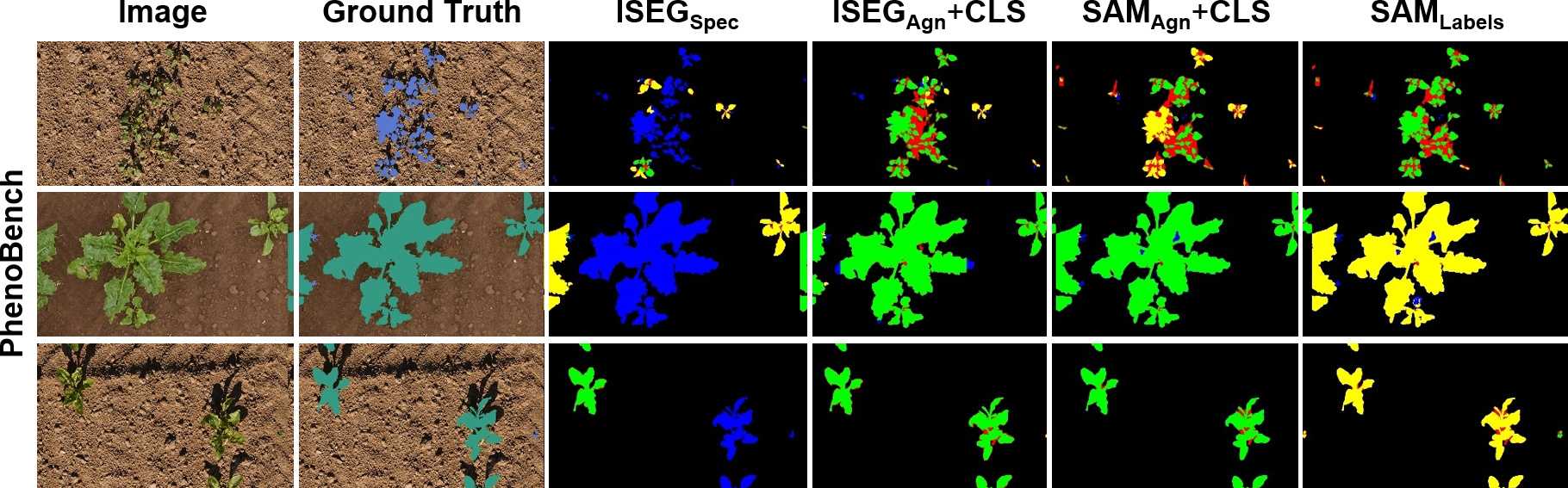}
    \caption{Representative results of unsupervised and weakly supervised upcycling methods on \mbox{\textbf{\textit{PhenoBench}}} \cite{WeMa24} images. Left to right: input images, ground truth, upcycling results per method (green: correctly detected mask areas, yellow: correctly located areas assigned to wrong classes, red: incorrectly detected areas, blue: undetected instances).}
    \label{fig:results_pb}
\end{figure}

\begin{figure}
    \centering
    \includegraphics[width=1\columnwidth]{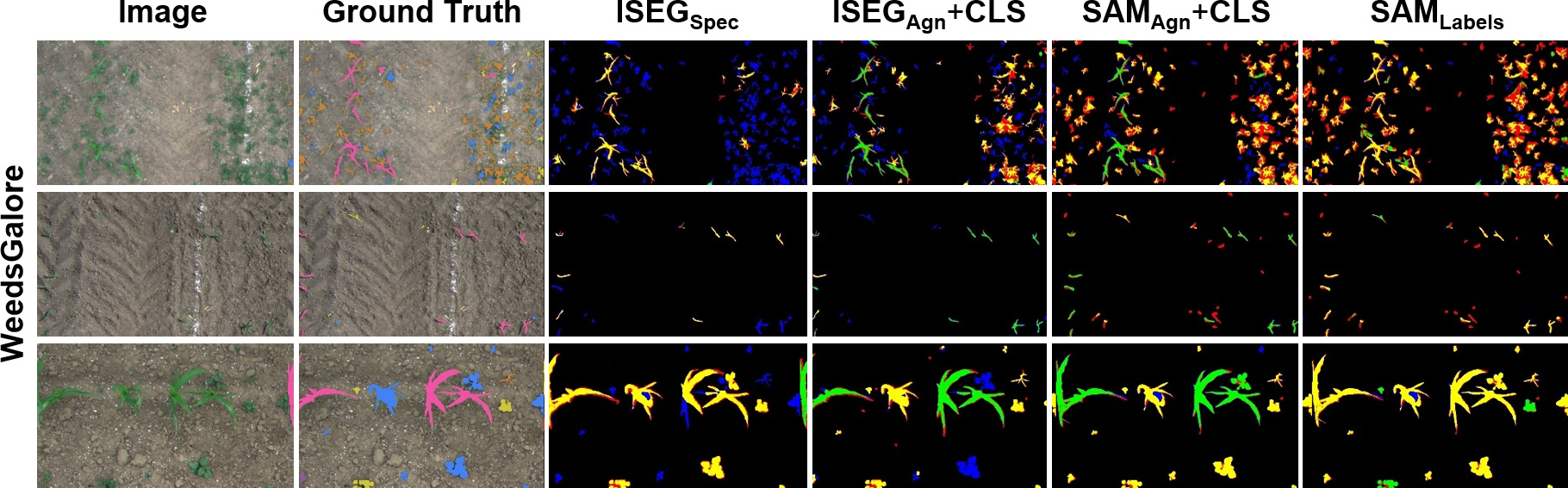}
    \caption{Representative results of unsupervised and weakly supervised upcycling methods on \textbf{\textit{WeedsGalore}} \cite{CeKu25} images. Left to right: input images, ground truth, upcycling results per method (green: correctly detected mask areas, yellow: correctly located areas assigned to wrong classes, red: incorrectly detected areas, blue: undetected instances).}
    \label{fig:results_wg}
\end{figure}

\begin{figure}
    \centering
    \includegraphics[width=1\columnwidth]{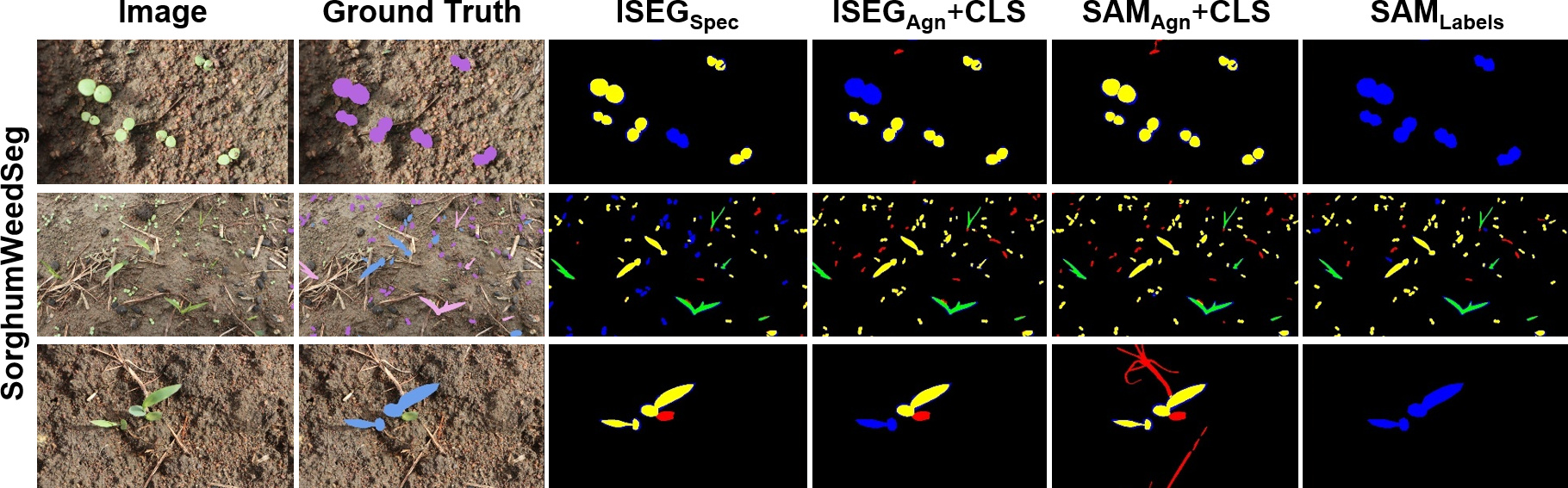}
    \caption{Representative results of unsupervised and weakly supervised upcycling methods on \textbf{\textit{SorghumWeedSeg}} \cite{JuTh24} images. Left to right: input images, ground truth, upcycling results per method (green: correctly detected mask areas, yellow: correctly located areas assigned to wrong classes, red: incorrectly detected areas, blue: undetected instances).}
    \label{fig:results_sw}
\end{figure}

\begin{figure}
    \centering
    \includegraphics[width=1\columnwidth]{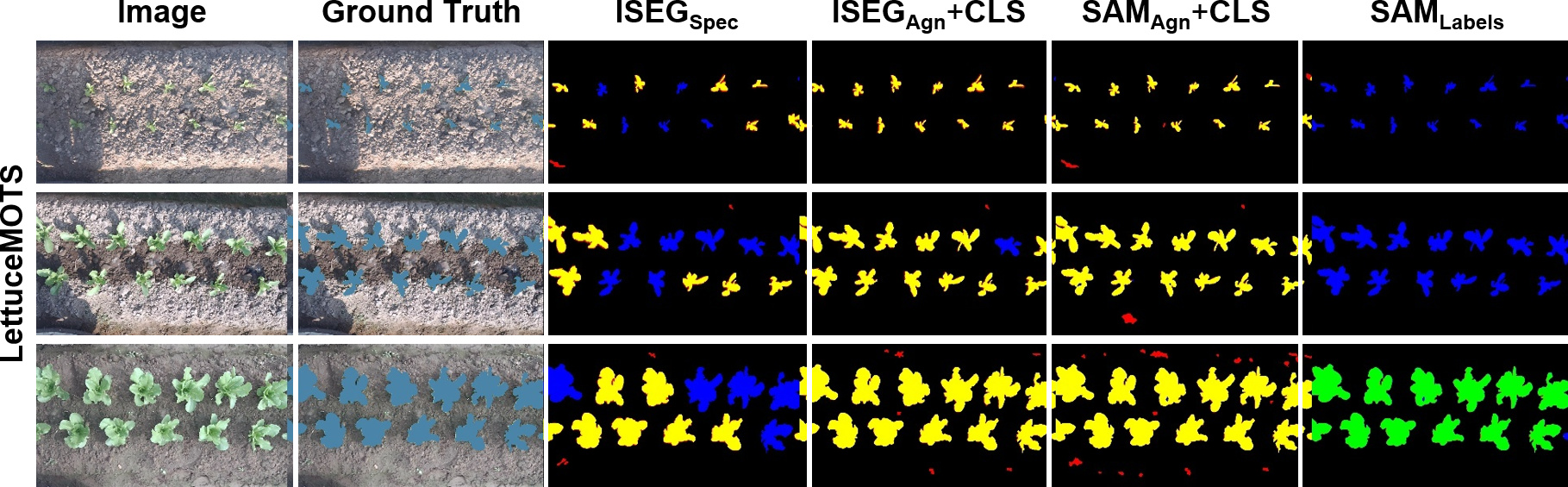}
    \caption{Representative results of unsupervised and weakly supervised upcycling methods on \mbox{\textbf{\textit{LettuceMOTS}}} \cite{HuSu24} images. Left to right: input images, ground truth, upcycling results per method (green: correctly detected mask areas, yellow: correctly located areas assigned to wrong classes, red: incorrectly detected areas, blue: undetected instances).}
    \label{fig:results_lm}
\end{figure}

\begin{figure}
    \centering
    \includegraphics[width=1\columnwidth]{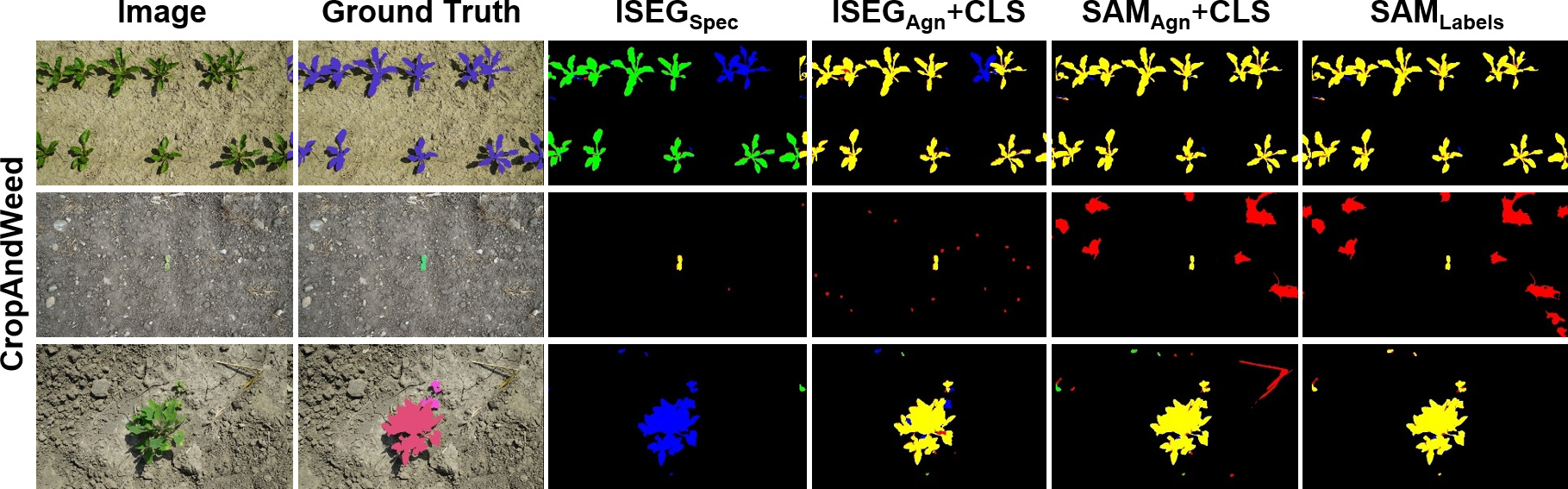}
    \caption{Representative results of unsupervised and weakly supervised upcycling methods on \textbf{\textit{CropAndWeed}} \cite{StTr23} images. Left to right: input images, ground truth, upcycling results per method (green: correctly detected mask areas, yellow: correctly located areas assigned to wrong classes, red: incorrectly detected areas, blue: undetected instances).}
    \label{fig:results_cw}
\end{figure}

The following statistics and visualizations complement the method comparison in \cref{sec:method_ablation} of the paper with more detailed quantitative and qualitative results. \cref{tab:results_agnostic_detail} provides recall and precision values, as well as average inference times, of class-agnostic approaches for each individual dataset, forming the basis for the mean values in \cref{tab:results_agnostic} of the main paper. To provide more insights into the composition and characteristics of each dataset, \cref{tab:class_results_pb,tab:class_results_wg,tab:class_results_sw,tab:class_results_lm,tab:class_results_cw} list mAP\textsuperscript{50-95} values for the mapped target classes of each dataset, which differ widely in their numbers and granularities. To visualize the corresponding differences in recording setups, \cref{fig:results_pb,fig:results_wg,fig:results_sw,fig:results_lm,fig:results_cw} show additional representative samples from each dataset, along with instance masks produced by unsupervised and weakly supervised methods.

\begin{table}
    \caption{Precision and recall of instance segmentation (at threshold with best F1 score), along with average inference \textbf{t}ime per image, on each target dataset (\textbf{P}heno\textbf{B}ench, \textbf{W}eeds\textbf{G}alore, \textbf{S}orghum\textbf{W}eedSeg, \textbf{L}ettuce\textbf{M}OTS, \textbf{C}ropAnd\textbf{W}eed) for agnostic methods. \textit{ISEG\textsubscript{Agn}} is a single-class instance-segmentation model trained on the remaining four datasets, \textit{SAM\textsubscript{Agn}} denotes results of SAM 3 prompted with generic superclass names. The baseline represents average and best results of individual models trained on the remaining four datasets for each target.}
    \label{tab:results_agnostic_detail}
    \centering
    \resizebox{1\linewidth}{!}{
    \begin{tabular}{@{}l@{\hspace{0.9em}}rrrrr@{\hspace{0.9em}}rrrrr@{\hspace{0.9em}}rrrrr@{}}
        \toprule
         & \multicolumn{5}{c}{\textbf{Precision}} & \multicolumn{5}{c}{\textbf{Recall}} & \multicolumn{5}{c}{\textbf{t [ms]}} \\
        & PB & WG & SW & LM & CW & PB & WG & SW & LM & CW & PB & WG & SW & LM & CW \\
        \midrule
        \textit{Bl\textsubscript{Overall}} & 68.4 & 79.0 & 53.4 & 87.8 & 56.9 & 22.4 & 11.0 & 42.5 & 60.8 & 37.9 & 38.4 & 45.5 & 35.7 & 31.0 & 22.4 \\
        \textit{Bl\textsubscript{Best}} & 91.2 & 85.7 & 66.0 & 95.0 & 77.4 & 34.3 & 19.2 & 48.5 & 90.1 & 60.1 & 37.5 & 40.4 & 29.3 & 28.6 & 21.0 \\
        \textit{ISEG\textsubscript{Agn}} & 89.8 & 71.3 & 72.0 & 97.0 & 64.7 & 61.1 & 22.7 & 70.3 & 93.0 & 57.2 & 46.1 & 78.4 & 51.0 & 38.6 & 26.8 \\
        \textit{SAM\textsubscript{Agn}} & 78.2 & 67.9 & 61.1 & 98.0 & 76.6 & 61.2 & 51.7 & 66.3 & 96.0 & 68.7 & 633.7 & 885.1 & 685.1 & 618.0 & 592.4 \\
        \bottomrule
    \end{tabular}
    }
\end{table}

\clearpage

\begin{table}
    \caption{Instance-mask accuracy (\% mAP\textsuperscript{50-95}) of unsupervised, weakly supervised and semi-supervised methods for each target class in the \textit{\textbf{PhenoBench}} \cite{WeMa24} dataset.}
    \label{tab:class_results_pb}
    \centering
    \resizebox{1\linewidth}{!}{
    \begin{tabular}{@{}l@{\quad}c@{\hspace{0.3em}}c@{\hspace{0.3em}}c@{\hspace{0.3em}}c@{\hspace{0.3em}}c@{\hspace{0.3em}}c@{\hspace{0.3em}}c@{\hspace{0.3em}}c@{}}
        \toprule
         \multirow{2}{*}{\textbf{PhenoBench}}
            & \multirow{2}{*}{\textbf{\textit{ISEG\textsubscript{Spec}}}} 
            & \textbf{\textit{ISEG\textsubscript{Agn}}}
            & \textbf{\textit{SAM\textsubscript{Agn}}}
            & \multirow{2}{*}{\textbf{\textit{SAM\textsubscript{Labels}}}} 
            & \textbf{\textit{ISEG\textsubscript{Agn}}}
            & \textbf{\textit{SAM\textsubscript{Agn}}}
            & \multirow{2}{*}{\textbf{\textit{SAM\textsubscript{Box}}}} 
            & \multirow{2}{*}{\textbf{\textit{SAM\textsubscript{Pt}}}} \\
        
            & & +\textit{\textbf{CLS}~~~} & +\textit{\textbf{CLS}~~~} & & +\textit{\textbf{Box~~~}} & +\textit{\textbf{Box~~~}} & & \\
        \midrule
        SugarBeet & 47.5 & 54.7 & 46.3 & 0.0 & 62.1 & 52.1 & 77.9 & 31.7 \\
        Weed & 16.3 & 30.0 & 25.1 & 9.0 & 40.0 & 35.7 & 56.6 & ~~1.6 \\        
        \bottomrule
    \end{tabular}
    }
\end{table}

\begin{table}
    \caption{Instance-mask accuracy (\% mAP\textsuperscript{50-95}) of unsupervised, weakly supervised and semi-supervised methods for each target class in the \textit{\textbf{WeedsGalore}} \cite{CeKu25} dataset.}
    \label{tab:class_results_wg}
    \centering
    \resizebox{1\linewidth}{!}{
    \begin{tabular}{@{}l@{\quad}c@{\hspace{0.3em}}c@{\hspace{0.3em}}c@{\hspace{0.3em}}c@{\hspace{0.3em}}c@{\hspace{0.3em}}c@{\hspace{0.3em}}c@{\hspace{0.3em}}c@{}}
        \toprule
        \multirow{2}{*}{\textbf{WeedsGalore}}
            & \multirow{2}{*}{\textbf{\textit{ISEG\textsubscript{Spec}}}} 
            & \textbf{\textit{ISEG\textsubscript{Agn}}}
            & \textbf{\textit{SAM\textsubscript{Agn}}}
            & \multirow{2}{*}{\textbf{\textit{SAM\textsubscript{Labels}}}} 
            & \textbf{\textit{ISEG\textsubscript{Agn}}}
            & \textbf{\textit{SAM\textsubscript{Agn}}}
            & \multirow{2}{*}{\textbf{\textit{SAM\textsubscript{Box}}}} 
            & \multirow{2}{*}{\textbf{\textit{SAM\textsubscript{Pt}}}} \\
        
            & & +\textit{\textbf{CLS}~~~} & +\textit{\textbf{CLS}~~~} & & +\textit{\textbf{Box~~~}} & +\textit{\textbf{Box~~~}} & & \\
        \midrule
        BarnyardGrass & ~~0.0 & ~~0.0 & ~~0.0 & 0.6 & 6.0 & 10.0 & 18.8 & ~~7.9 \\
        Maize & 26.0 & 16.9 & 17.0 & 0.0 & 22.1 & 25.2 & 35.7 & ~~1.2 \\
        PotatoWeed & ~~0.0 & ~~1.0 & ~~0.0 & 0.8 & 24.3 & 36.1 & 56.8 & ~~2.1 \\
        RedRootAmaranth & 26.2 & 16.0 & 20.8 & 0.0 & 21.5 & 33.5 & 51.6 & 17.1 \\
        Weed & ~~1.0 & ~~1.6 & ~~1.8 & 1.9 & 16.6 & 20.7 & 40.0 & ~~8.3 \\        
        \bottomrule
    \end{tabular}
    }
\end{table}

\begin{table}
    \caption{Instance-mask accuracy (\% mAP\textsuperscript{50-95}) of unsupervised, weakly supervised and semi-supervised methods for each target class in the \textit{\textbf{SorghumWeedSeg}} \cite{JuTh24} dataset.}
    \label{tab:class_results_sw}
    \centering
    \resizebox{1\linewidth}{!}{
    \begin{tabular}{@{}l@{\quad}c@{\hspace{0.3em}}c@{\hspace{0.3em}}c@{\hspace{0.3em}}c@{\hspace{0.3em}}c@{\hspace{0.3em}}c@{\hspace{0.3em}}c@{\hspace{0.3em}}c@{}}
        \toprule
        \multirow{2}{*}{\textbf{SorghumWeedSeg}}
            & \multirow{2}{*}{\textbf{\textit{ISEG\textsubscript{Spec}}}} 
            & \textbf{\textit{ISEG\textsubscript{Agn}}}
            & \textbf{\textit{SAM\textsubscript{Agn}}}
            & \multirow{2}{*}{\textbf{\textit{SAM\textsubscript{Labels}}}} 
            & \textbf{\textit{ISEG\textsubscript{Agn}}}
            & \textbf{\textit{SAM\textsubscript{Agn}}}
            & \multirow{2}{*}{\textbf{\textit{SAM\textsubscript{Box}}}} 
            & \multirow{2}{*}{\textbf{\textit{SAM\textsubscript{Pt}}}} \\
        
            & & +\textit{\textbf{CLS}~~~} & +\textit{\textbf{CLS}~~~} & & +\textit{\textbf{Box~~~}} & +\textit{\textbf{Box~~~}} & & \\
        \midrule
        Dicots & 0.0 & 0.0 & 0.0 & 0.0 & 29.3 & 22.4 & 39.7 & 9.0 \\
        Grasses & 8.1 & 5.4 & 3.5 & 7.1 & 21.3 & 15.7 & 21.7 & 0.3 \\
        Sorghum & 0.0 & 3.8 & 0.0 & 0.0 & 23.1 & 18.6 & 36.3 & 0.3 \\        
        \bottomrule
    \end{tabular}
    }
\end{table}

\begin{table}
    \caption{Instance-mask accuracy (\% mAP\textsuperscript{50-95}) of unsupervised, weakly supervised and semi-supervised methods for each target class in the \textit{\textbf{LettuceMOTS}} \cite{HuSu24} dataset.}
    \label{tab:class_results_lm}
    \centering
    \resizebox{1\linewidth}{!}{
    \begin{tabular}{@{}l@{\quad}c@{\hspace{0.3em}}c@{\hspace{0.3em}}c@{\hspace{0.3em}}c@{\hspace{0.3em}}c@{\hspace{0.3em}}c@{\hspace{0.3em}}c@{\hspace{0.3em}}c@{}}
        \toprule
        \multirow{2}{*}{\textbf{LettuceMOTS}}
            & \multirow{2}{*}{\textbf{\textit{ISEG\textsubscript{Spec}}}} 
            & \textbf{\textit{ISEG\textsubscript{Agn}}}
            & \textbf{\textit{SAM\textsubscript{Agn}}}
            & \multirow{2}{*}{\textbf{\textit{SAM\textsubscript{Labels}}}} 
            & \textbf{\textit{ISEG\textsubscript{Agn}}}
            & \textbf{\textit{SAM\textsubscript{Agn}}}
            & \multirow{2}{*}{\textbf{\textit{SAM\textsubscript{Box}}}} 
            & \multirow{2}{*}{\textbf{\textit{SAM\textsubscript{Pt}}}} \\
        
            & & +\textit{\textbf{CLS}~~~} & +\textit{\textbf{CLS}~~~} & & +\textit{\textbf{Box~~~}} & +\textit{\textbf{Box~~~}} & & \\
        \midrule
        Lettuce & 0.0 & 0.0 & 0.0 & 57.6 & 87.8 & 86.8 & 89.4 & 84.9 \\
        \bottomrule
    \end{tabular}
    }
\end{table}

\begin{table}
    \caption{Instance-mask accuracy (\% mAP\textsuperscript{50-95}) of unsupervised, weakly supervised and semi-supervised methods for each target class in the \textit{\textbf{CropAndWeed}} \cite{StTr23} dataset.}
    \label{tab:class_results_cw}
    \centering
    \resizebox{1\linewidth}{!}{
    \begin{tabular}{@{}l@{\quad}c@{\hspace{0.3em}}c@{\hspace{0.3em}}c@{\hspace{0.3em}}c@{\hspace{0.3em}}c@{\hspace{0.3em}}c@{\hspace{0.3em}}c@{\hspace{0.3em}}c@{}}
        \toprule
        \multirow{2}{*}{\textbf{CropAndWeed}}
            & \multirow{2}{*}{\textbf{\textit{ISEG\textsubscript{Spec}}}} 
            & \textbf{\textit{ISEG\textsubscript{Agn}}}
            & \textbf{\textit{SAM\textsubscript{Agn}}}
            & \multirow{2}{*}{\textbf{\textit{SAM\textsubscript{Labels}}}} 
            & \textbf{\textit{ISEG\textsubscript{Agn}}}
            & \textbf{\textit{SAM\textsubscript{Agn}}}
            & \multirow{2}{*}{\textbf{\textit{SAM\textsubscript{Box}}}} 
            & \multirow{2}{*}{\textbf{\textit{SAM\textsubscript{Pt}}}} \\
        
            & & +\textit{\textbf{CLS}~~~} & +\textit{\textbf{CLS}~~~} & & +\textit{\textbf{Box~~~}} & +\textit{\textbf{Box~~~}} & & \\
            \midrule
            AnnualMeadowGrass & ~~0.0 & 0.0 & 0.0 & 0.0 & ~~4.4 & ~~6.8 & 16.3 & ~~5.3 \\
            AnnualMercury & ~~0.0 & 0.0 & 0.0 & 0.0 & 42.6 & 50.0 & 50.5 & ~~1.1 \\
            BarnyardGrass & ~~0.2 & 0.0 & 0.0 & 2.3 & 20.4 & 23.0 & 36.3 & ~~0.4 \\
            BlackBindweed & ~~0.0 & 0.0 & 0.0 & 0.0 & 35.9 & 43.5 & 53.6 & 11.4 \\
            BroadBean & ~~0.0 & 0.0 & 0.0 & 0.0 & 66.4 & 66.1 & 71.9 & 39.8 \\
            CharlockMustard & ~~0.0 & 0.0 & 0.0 & 0.0 & 16.6 & 45.7 & 54.4 & 10.6 \\
            CommonBuckwheat & ~~0.0 & 0.0 & 0.0 & 0.0 & 44.5 & 60.8 & 60.5 & 20.1 \\
            CommonChickweed & ~~0.0 & 0.0 & 0.0 & 0.1 & 22.2 & 40.6 & 50.5 & 12.2 \\
            CommonCorncockle & ~~0.0 & 3.9 & 2.2 & 0.0 & 52.5 & 46.0 & 55.0 & 18.8 \\
            CommonFumitory & ~~0.0 & 0.0 & 0.0 & 0.0 & 32.3 & 57.1 & 63.3 & 25.9 \\
            CommonKnotgrass & ~~0.0 & 0.0 & 0.0 & 0.0 & 13.8 & 21.7 & 31.1 & ~~3.4 \\
            CommonPoppy & ~~0.0 & 0.0 & 0.0 & 0.0 & 39.2 & 50.2 & 56.6 & 21.5 \\
            CommonSunflower & ~~0.0 & 0.1 & 0.1 & 0.0 & 72.7 & 62.8 & 70.0 & 44.7 \\
            CommonWildOat & ~~0.0 & 0.6 & 0.5 & 0.0 & 23.5 & 26.0 & 35.7 & ~~8.7 \\
            CopseBindweed & ~~0.0 & 0.0 & 0.0 & 0.0 & 29.2 & 44.1 & 47.3 & 19.7 \\
            CornSpurry & ~~0.0 & 0.0 & 0.1 & 0.0 & 18.7 & 26.5 & 34.9 & ~~8.1 \\
            Cornflower & ~~0.0 & 0.0 & 0.1 & 0.0 & 33.1 & 38.7 & 48.9 & 10.8 \\
            CreepingThistle & ~~0.0 & 0.3 & 0.6 & 0.0 & 50.9 & 52.2 & 60.9 & 24.9 \\
            FieldBindweed & ~~0.0 & 0.2 & 0.8 & 0.0 & 16.9 & 45.5 & 51.8 & 14.5 \\
            FieldPennycress & ~~0.0 & 0.4 & 0.2 & 0.0 & 60.8 & 62.9 & 66.8 & 42.0 \\
            FieldSowthistle & ~~0.0 & 0.0 & 0.0 & 0.1 & 31.4 & 44.3 & 55.2 & 15.6 \\
            Geraniums & ~~0.0 & 0.0 & 0.0 & 0.1 & 13.8 & 23.0 & 43.4 & ~~8.1 \\
            Grasses & ~~0.0 & 0.0 & 0.0 & 0.0 & ~~5.3 & ~~4.1 & ~~9.3 & ~~2.3 \\
            GreenFoxtail & ~~0.0 & 0.0 & 0.0 & 0.0 & 15.5 & 15.8 & 23.5 & ~~3.1 \\
            HairyCrabgrass & ~~0.0 & 0.0 & 0.0 & 0.7 & 11.2 & 39.8 & 45.0 & 13.6 \\
            HedgeMustard & ~~0.0 & 0.0 & 0.0 & 0.0 & 10.2 & 38.3 & 50.1 & 13.9 \\
            IvyLeavedSpeedwell & ~~0.0 & 0.0 & 0.0 & 0.0 & 36.7 & 45.9 & 53.3 & 21.6 \\
            Maize & 10.1 & 0.0 & 0.0 & 0.0 & 57.0 & 52.8 & 61.9 & 30.5 \\
            MapleLeafGoosefoot & ~~0.0 & 0.0 & 0.0 & 0.0 & 55.3 & 66.1 & 73.6 & 33.9 \\
            MeadowGrasses & ~~0.0 & 0.0 & 0.0 & 0.9 & 14.8 & 18.0 & 23.9 & ~~4.1 \\
            MouseBarley & ~~0.0 & 0.0 & 0.0 & 0.0 & 19.6 & 26.4 & 31.5 & ~~6.7 \\
            Pea & ~~0.0 & 0.0 & 0.0 & 0.0 & 30.8 & 38.7 & 49.9 & 24.6 \\
            Peppermint & ~~0.0 & 0.0 & 0.0 & 0.0 & 30.2 & 47.1 & 49.5 & 11.4 \\
            Pineappleweed & ~~0.0 & 0.0 & 0.0 & 0.0 & ~~8.4 & 30.0 & 37.6 & ~~7.9 \\
            Potato & ~~0.0 & 0.0 & 0.0 & 0.0 & 47.9 & 66.2 & 82.0 & 25.7 \\
            PotatoWeed & ~~0.0 & 0.0 & 0.0 & 0.3 & 38.8 & 47.5 & 59.8 & 22.9 \\
            RedRootAmaranth & ~~0.0 & 0.0 & 0.0 & 0.0 & 34.3 & 37.1 & 53.5 & 10.2 \\
            Redshank & ~~0.0 & 0.0 & 0.0 & 0.0 & 29.3 & 39.6 & 47.7 & 14.0 \\
            RibwortPlantain & ~~0.0 & 0.0 & 0.0 & 0.0 & 32.0 & 27.5 & 37.5 & ~~6.6 \\
            RyeBrome & ~~0.0 & 0.0 & 0.0 & 0.0 & 21.3 & 23.7 & 30.9 & ~~4.0 \\
            SmallGeranium & ~~0.0 & 0.0 & 0.0 & 0.0 & ~~6.8 & 22.7 & 40.2 & ~~4.8 \\
            SoftBrome & ~~0.0 & 0.0 & 0.0 & 0.0 & 12.6 & 19.0 & 24.3 & ~~5.4 \\
            Soybean & ~~0.0 & 0.0 & 0.0 & 0.0 & 31.2 & 40.8 & 63.7 & 14.5 \\
            Squash & ~~0.0 & 0.0 & 0.0 & 0.0 & 73.6 & 75.3 & 82.1 & 50.7 \\
            SugarBeet & 34.8 & 0.0 & 0.0 & 0.0 & 60.6 & 55.5 & 64.9 & 21.0 \\
            Thornapple & ~~0.0 & 0.0 & 0.0 & 0.0 & 57.2 & 48.3 & 64.1 & 23.1 \\
            Vegetation & ~~1.6 & 1.6 & 2.9 & 1.5 & 18.9 & 28.5 & 42.6 & 11.3 \\
            WhiteGoosefoot & ~~0.0 & 0.0 & 0.0 & 0.0 & 33.4 & 47.3 & 56.9 & 16.7 \\
            WildChamomile & ~~0.0 & 0.0 & 0.0 & 0.0 & 20.1 & 43.0 & 55.3 & 20.2 \\
            WildPansy & ~~0.0 & 0.0 & 0.0 & 0.0 & 36.5 & 50.9 & 56.1 & 22.2 \\
            Wintercress & ~~0.0 & 0.0 & 0.0 & 0.0 & 37.1 & 47.5 & 59.3 & 21.3 \\
            \bottomrule
        \end{tabular}
    }
\end{table}

\end{document}